\documentclass[lettersize,journal]{IEEEtran}
\usepackage{amsmath,amsfonts}
\usepackage{array}
\usepackage[caption=false,font=normalsize,labelfont=sf,textfont=sf]{subfig}
\usepackage{textcomp}
\usepackage{stfloats}
\usepackage{url}
\usepackage{verbatim}
\usepackage{graphicx}
\usepackage{cite}
\def \ie{{\em i.e.}}
\def \eg{{\em e.g.}}

\usepackage[linesnumbered,ruled]{algorithm2e}
\usepackage{verbatim}
\usepackage{bm}
\usepackage{epsfig,latexsym}
\usepackage{float}
\usepackage{indentfirst}
\usepackage{times}
\usepackage{psfrag}
\usepackage{multicol}
\usepackage{multirow}
\usepackage{stfloats}
\usepackage{url}
\usepackage{threeparttable}
\usepackage{makecell}
\usepackage{arydshln}
\usepackage{pifont}   
\usepackage{bbding}
\usepackage{rotating}
\usepackage{bbm}
\usepackage{colortbl}
\usepackage{array}
\usepackage{tabulary}
\definecolor{mygray}{gray}{.9}
\usepackage{hyperref}

\newcommand{\OPBL}{\textcolor{blue}}

\usepackage[normalem]{ulem}
\begin{document}

\title{Arbitrary Ratio Feature Compression via Next Token Prediction}

\author{Yufan Liu,
    Daoyuan Ren, 
    Zhipeng Zhang,
    Wenyang Luo,
	Bing Li,
	Weiming Hu,
	Stephen Maybank,~\IEEEmembership{Fellow,~IEEE}
\thanks{Corresponding Author: Bing Li.}
\thanks{Yufan Liu, Daoyuan Ren, Wenyang Luo, Bing Li, and Weiming Hu are with State Key Laboratory of Multimodal Artificial Intelligence Systems, Institution of Automation, Chinese Academy of Sciences; School of Artificial Intelligence, University of Chinese Academy of Sciences, and CAS Center for Excellence in Brain Science and Intelligence Technology. Bing Li is also with People AI, Inc. (e-mail: \{yufan.liu, rendaoyuan2023, luowenyang2020\}@ia.ac.cn;\{bli, wmhu\}@nlpr.ia.ac.cn).}
\thanks{Zhipeng Zhang is with the School of Artificial Intelligence, Shanghai Jiao Tong University. (e-mail: zhipeng.zhang.cv@outlook.com)}
\thanks{Stephen Maybank is an emeritus professor in the School of Computer Science and Mathematics, Birkbeck College, University of London, London WC1E 7HX, U.K. (e-mail: sjmaybank@gmail.com).}
\thanks{Manuscript received Augest 26, 2025}}

\markboth{Journal of \LaTeX\ Class Files,~Vol.~14, No.~8, August~2021}%
{Liu \MakeLowercase{\textit{et al.}}: Arbitrary Ratio Feature Compression via Next Token Prediction}


\maketitle

\begin{abstract}
Feature compression is increasingly important for improving the efficiency of downstream tasks, especially in applications involving large-scale or multi-modal data. While existing methods typically rely on dedicated models for achieving specific compression ratios, they are often limited in flexibility and generalization. In particular, retraining is necessary when adapting to a new compression ratio.
To address this limitation, we propose a novel and flexible Arbitrary Ratio Feature Compression (ARFC) framework, which supports any compression ratio with a single model, eliminating the need for multiple specialized models. 
At its core, the Arbitrary Ratio Compressor (ARC) is an auto-regressive model that performs compression via next-token prediction. This allows the compression ratio to be controlled at inference simply by adjusting the number of generated tokens. 
To enhance the quality of the compressed features, two key modules are introduced. The Mixture of Solutions (MoS) module refines the compressed tokens by utilizing multiple compression results (solutions), reducing uncertainty and improving robustness. 
The Entity Relation Graph Constraint (ERGC) is integrated into the training process to preserve semantic and structural relationships during compression.
Extensive experiments on cross-modal retrieval, image classification, and image retrieval tasks across multiple datasets demonstrate that our method consistently outperforms existing approaches at various compression ratios. Notably, in some cases, it even surpasses the performance of the original, uncompressed features. These results validate the effectiveness and versatility of ARFC for practical, resource-constrained scenarios.
\end{abstract}


\section{Introduction}\label{sec1}
The rapid development of deep learning and artificial intelligence has led to improved feature representation~\cite{ping2013review}. Modern models can extract rich, high-level semantic features from raw data such as images and texts, capturing complex patterns and enabling strong performance across a wide range of downstream tasks, including classification~\cite{lu2007survey}, retrieval~\cite{wang2016comprehensive}, and clustering~\cite{saxena2017review}.

\begin{figure}[htbp]
  \centering
  \includegraphics[width=1.01\linewidth]{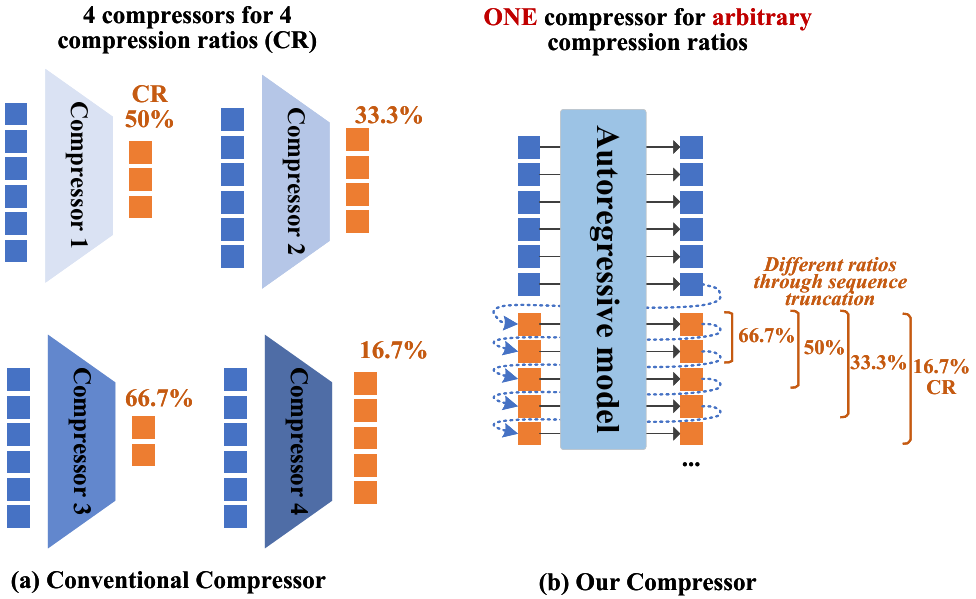}
  \caption{Comparison between conventional compressors and our proposed compressor. The conventional feature compression methods require training multiple compressors for various compression ratios separately, while our method only needs to train ONE compressor to support arbitrary compression ratios.}
  \label{fig:fig1} 
\end{figure}

The emergence of vision-language pre-trained models like CLIP~\cite{radford2021learning} further enhances feature expressiveness by encoding multi-modal data into a shared embedding space. This allows for direct comparison between different modalities (\eg, image and text), enabling powerful applications such as zero-shot recognition~\cite{radford2021learning} and cross-modal retrieval~\cite{wang2016comprehensive}. However, as these models become more expressive, the dimension of the extracted features tends to increase, posing challenges to storage, transmission, and computational efficiency in downstream tasks.

To address this issue, feature compression~\cite{jia2022feature} has gained increasing attention. The use of compression is to reduce feature dimensions while preserving feature utility, allowing compressed features to perform nearly as well as the original ones in downstream tasks. Existing methods typically adopt an autoencoder-like structure: a compressor encodes the input feature into a lower-dimensional representation, and a decoder reconstructs the original feature to supervise training. Most approaches require separate models for different compression ratios, which is inefficient and impractical when facing dynamic resource constraints.

In contrast, we propose a novel Arbitrary Ratio Feature Compression (ARFC) framework that achieves arbitrary-ratio feature compression with only one training stage. Unlike conventional methods, our approach does not rely on multiple compressors for different compression ratios. Instead, it leverages a next token prediction mechanism, inspired by Auto-Regressive (AR) modeling~\cite{dickey1979distribution}, to enable flexible compression at any desired ratio, as illustrated in Fig. \ref{fig:fig1}. 
Specifically, our framework consists of three key components: Arbitrary Ratio Compressor (ARC) via Next-Token Prediction, Mixture of Solutions (MoS), and Entity Relation Graph Constraint (ERGC). 
ARC enables auto-regressive style compression, where the feature vector is processed sequentially, allowing the model to dynamically adjust the compression ratio without retraining. MoS refines the compressed feature by aggregating multiple basic solutions. ERGC is a novel regularization term that preserves relationships among entities during compression. It ensures that both the semantic and geometric structures of the original feature space are maintained in the compressed space. It is applied consistently across both ARC and MoS modules.

Extensive experiments demonstrate that our method achieves superior performance across a wide range of datasets and tasks, including cross-modal retrieval~\cite{wang2016comprehensive}, image retrieval~\cite{dubey2021decade}, and image classification~\cite{rawat2017deep}, under various compression ratios. Our approach outperforms a wide range of competing methods, including Querying Transformer (Q-Former)~\cite{li2023blip}, autoencoders~\cite{pinheiro2021variational}, and Post-Training Quantization (PTQ)~\cite{liu2021post}. Specifically, our method even achieves up to 75\% lossless compression in most scenarios.

The main contributions of this work are summarized as follows:
\begin{itemize}
\item We propose a novel Arbitrary Ratio Feature Compression (ARFC) framework that requires the training of one model only. The core component Arbitrary Ratio Compressor (ARC) leverages a next token prediction mechanism to enable flexible compression at any desired ratio.
\item We introduce Mixture of Solutions (MoS) blocks and Entity Relation Graph Constraint (ERGC) to enhance feature quality and maintain structural consistency, ensuring robustness across different compression ratios.
\item We conduct comprehensive experiments across multiple datasets and tasks, showing that our method consistently outperforms existing feature compression methods, particularly in terms of flexibility and performance retention under varying compression ratios. 
\end{itemize}

\section{Related Work}\label{sec2}
Feature compression aims to reduce the dimensions of the features while preserving essential information. Unlike model compression~\cite{liu2022learning,liu2019knowledge,ruan2020edp}, which reduces model size, feature compression enhances efficiency in storage, transmission, and downstream tasks, especially under limited computational resources or bandwidth.

Related researches on feature compression are categorized into multiple perspectives. 
Sec.~\ref{sec:classic} reviews classical dimension reduction methods (\eg, Principal Component Analysis (PCA)) which use linear or nonlinear transformations to project data into lower-dimensional spaces. 
Sec.~\ref{sec:neural_encoders} discusses learned feature compression via neural encoders (\eg, autoencoders and Q-Former) which learn compact and semantically rich representations through end-to-end training.
Sec.~\ref{sec:quantization} focuses on quantization-based compression, which reduces precision by converting floating-point features and weights into low-bit representations.

\subsection{Classical Dimension Reduction}\label{sec:classic}
Classical dimension reduction techniques aim to reduce the feature space while retaining essential information. Principal Component Analysis (PCA)~\cite{abdi2010principal} is a widely used linear method that projects data onto a lower-dimensional space by maximizing variance. Other classical methods include Linear Discriminant Analysis (LDA)~\cite{martinez2001pca}, which seeks directions that maximize class separability, and t-SNE~\cite{maaten2008visualizing}, which focuses on preserving local structures in high-dimensional data. Isomap~\cite{tenenbaum2000global} extends PCA by considering geodesic distances between points, making it suitable for non-linear data.

Several techniques are commonly used for compressing image data, such as wavelet transforms and vector quantization. For example, Discrete Wavelet Transform combined with Vector Quantization (DWT-VQ)~\cite{patel2013image} compresses images while maintaining quality. These traditional methods are effective for single-modality data but often struggle with multimodal data and varying compression requirements.

\subsection{Learned Feature Compression via Neural Encoders}\label{sec:neural_encoders}
Neural network-based encoders offer a more flexible and powerful approach to feature compression by learning data-driven representations, such as autoencoders (AEs)~\cite{wang2014generalized} and variational autoencoders (VAEs) \cite{pinheiro2021variational}. A typical autoencoder contains an encoder to compress the input and a decoder to reconstruct the input. This framework enables nonlinear compression and has been widely applied in image and signal processing.

Convolutional autoencoders extend this idea to image data by exploiting spatial hierarchies through convolutional layers, making them especially effective for visual feature compression~\cite{cheng2018deep}. Variants such as Stacked Contractive Auto-encoders (SCAE) are used in LDFA~\cite{zhang2018local}, which learns local deep features from neighborhood structures and aligns them globally via affine transformations, preserving both local and global data characteristics.

In multimodal learning, learned compression plays a crucial role in bridging the gap between vision and language. BLIP-2~\cite{li2023blip} introduces a Querying Transformer (Q-Former) as a lightweight cross-modal projector that compresses high-dimensional visual features into compact semantic vectors. This design creates an information bottleneck while maintaining compatibility with large language models, achieving strong alignment with few trainable parameters.
The DeCo framework~\cite{yao2024deco} addresses issues related to double abstraction by decoupling pure token compression from semantic abstraction, using parameter-free 2D adaptive average pooling to preserve spatial locality and simplify cross-modal alignment.


Despite their power, learned encoders often require retraining for different compression ratios and are typically designed for single-modality data, limiting their flexibility and generalization.

\subsection{Quantization-based Compression}\label{sec:quantization}
Quantization algorithms aim to quantize the deep learning model to enable fixed-point computation and less memory space. For most cases, the output, such as the output feature vector, is also quantized. At this end, quantization is another way of achieving feature compression. Quantization methods are typically categorized into Quantization-Aware Training (QAT), which fine-tunes models during training, and Post-Training Quantization (PTQ)~\cite{liu2021post}, which requires no retraining and is more practical for real-world applications. Due to its low computational overhead, PTQ has gained increasing attention~\cite{zhao2023post}.

MinMax quantization~\cite{schuster1999review} is a simple and widely used method that reduces model size and computation by mapping floating-point values to low-bit integers using the data’s minimum and maximum values. 
Recent works improve PTQ through better error control. For example, QDROP~\cite{wei2022qdrop} applies dropout during calibration to improve activation quantization, while BRecQ~\cite{li2021brecq} introduces a block-wise reconstruction to preserve layer-wise feature distributions, going beyond whole-network output matching.

For Vision Transformers, specialized PTQ methods have been developed. 
\cite{liu2021post} introduce a ranking loss to preserve the relative order of self-attention outputs after quantization, maintaining model functionality.  
PTQ4ViT~\cite{yuan2022ptq4vit} addresses poor activation calibration after Softmax and GELU~\cite{hendrycks2016gaussian} by proposing dual uniform quantization and Hessian-guided metrics for more accurate, efficient calibration. 
FQ-ViT~\cite{lin2021fq} tackles large channel variations in LayerNorm and skewed attention maps with Power-of-Two Factor (PTF) and Log Int Softmax (LIS), improving quantization precision and enabling efficient integer inference. 
More recently, \cite{sun2024p4q} propose Prompt for Quantization (P4Q), a method tailored for large Vision-Language Models (VLMs). P4Q introduces lightweight prompts and a low-bit adapter (QAdapter) to enable efficient quantization while maintaining performance, offering a promising path for deploying VLMs on resource-limited devices.

Despite these advances, quantization inevitably introduces numerical errors in operations like addition and multiplication. These errors propagate across layers, and in deep Transformer-based models, they can accumulate significantly, leading to performance degradation, especially in high-entropy modalities like vision, where information loss can distort cross-modal representations.

\begin{figure*}
  \centering
  \includegraphics[width=1\linewidth]{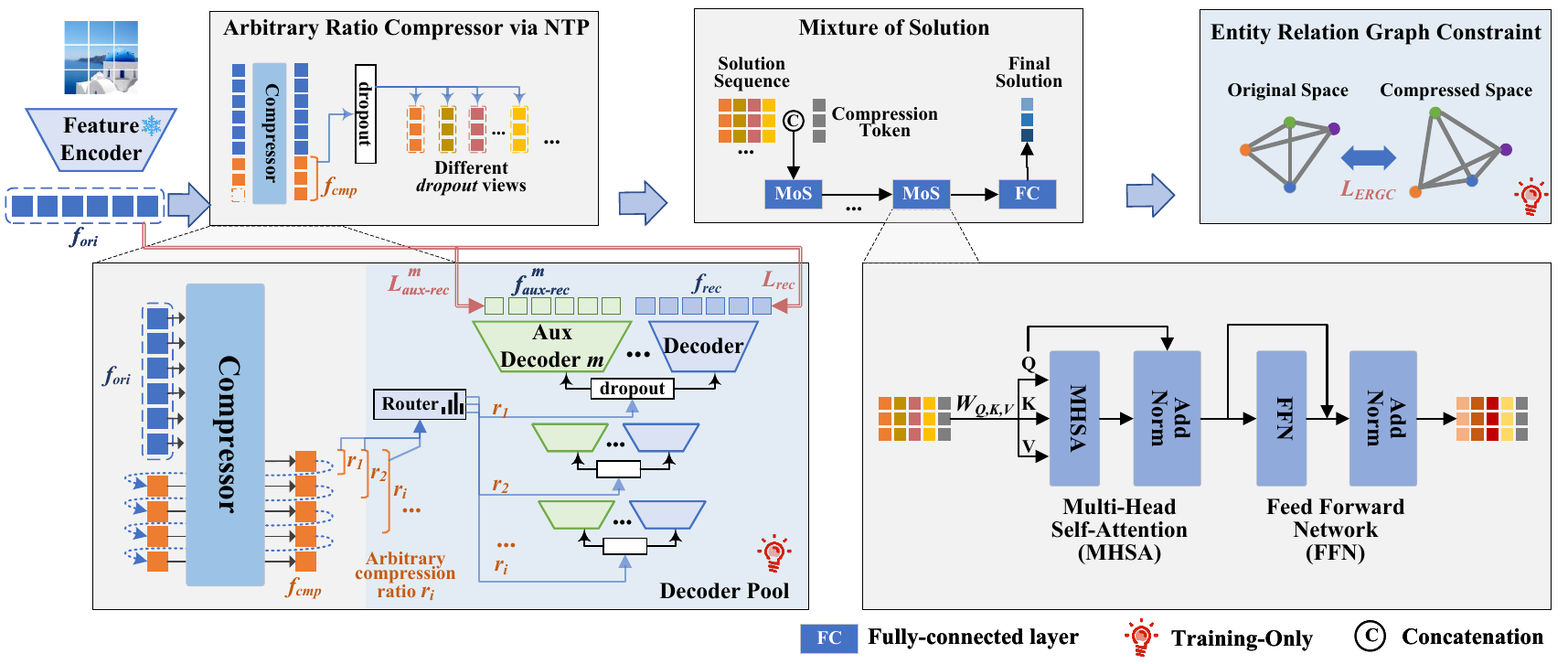}
  \caption{The overall framework of the proposed method. It consists of three key components, including the Arbitrary Ratio Compressor (ARC) via Next Token Prediction (NTP), the Mixture of Solution (MoS) blocks, and the Entity Relation Graph Constraint (ERGC).}
  \label{fig:framework} 
\end{figure*}

\section{Approach}
The overall framework of Arbitrary Ratio Feature Compression (ARFC) is illustrated in Fig. \ref{fig:framework}. It consists of three key components: Arbitrary Ratio Compressor (ARC) via Next Token Prediction (NTP) (introduced in Sec. \ref{sec:ARC}), the Mixture of Solution (MoS) blocks (described in Sec. \ref{sec:MoS}), and the Entity Relation Graph Constraint (ERGC) (introduced in Sec. \ref{sec:ERGC}). ARC, as the core module of ARFC, employs an auto-regressive approach to predict the compressed features in a token-by-token manner. By selecting an arbitrary number of output tokens, any prespecified Compression Ratio (CR) can be achieved. 
During training, these compressed features are reconstructed by decoders: a main decoder and several auxiliary decoders which ensure robust feature reconstruction and reduce the uncertainty of the ARC. 
A given input sample may yield a range of compressed results from the ARC. Each compressed result is treated as a basic solution. MoS is proposed to generate a better solution based on different basic solutions. Further, ERGC is proposed to preserve semantic relationships between multi-modal samples via a graph-based constraint. Each component is described in the following subsections. The final loss function and the optimization are introduced in Sec. \ref{sec:optimization}.

\begin{figure}
  \centering
  \includegraphics[width=0.95\linewidth]{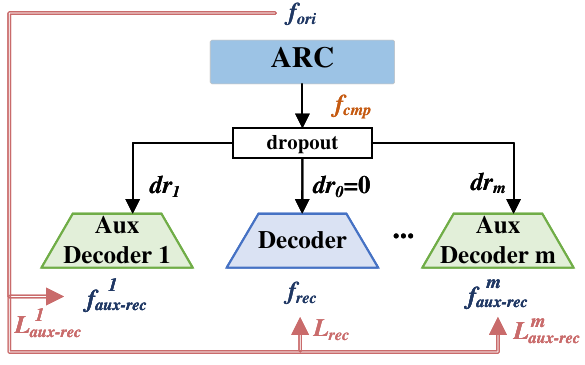}
  \caption{An example of an atom compressor with two auxiliary decoders. It consists of an encoder, a decoder, and several auxiliary decoders. The auxiliary decoders receive multi-view compressed features with different dropout rates and reconstruct the original feature.}
  \label{fig:AuxDecoder} 
\end{figure}

\subsection{Arbitrary Ratio Compressor via NTP}\label{sec:ARC}
Existing compression methods only support fixed CR compression. They require retraining a new encoder to support a new CR, leading to limited flexibility and high complexity. In contrast, the proposed ARFC enables compression at any desired ratio, using a single model.

\textbf{(1) Arbitrary ratio compression via next token prediction.}
Given an input (image or text), the feature $f_{ori}\in \mathbb{R}^D$ is extracted via a frozen encoder such as CLIP. Then $f_{ori}$ is partitioned into a sequence of $T$ tokens: 
\begin{equation}
\begin{aligned}
\label{equation:tokenization}
f_{ori} = (z_1, z_2, ..., z_T), \quad z_t\in \mathbb{R}^d.
\end{aligned}
\end{equation}
The token $z_t$ represents a local feature patch in a flattened form, with dimension $d=D/T$. 
A basic solution $f_{cmp}$ (\ie, the compressed feature at the least compression ratio $r_{min}=0$) has the same dimension as $z$: 
\begin{equation}
\begin{aligned}
\label{equation:solution}
f_{cmp} = (z_{T+1}, z_{T+2}, ..., z_{2T}), \quad z_t\in \mathbb{R}^d.
\end{aligned}
\end{equation}
Then, the feature compression task can be formulated as a next-token-prediction problem, where each token is predicted based on all previous tokens:
\begin{equation}
\begin{aligned}
\label{equation:probability}
&p(z_{T+1}, z_{T+2}, ..., z_{2T}|z_1, z_2, ..., z_T)\\
&=\prod_{t=T+1}^{2T} p_{\theta}(z_t|z_1, z_2, ..., z_{t-1}).
\end{aligned}
\end{equation}
Here, $p_{\theta}$ denotes a transformer-based auto-regressive model parameterized by $\theta$. It models the conditional distribution of the next token given the prefix history. During training, the model learns to reconstruct each token $z_t$ from its preceding context $(z_1, z_2, ..., z_{t-1})$. 
Then $r_{i}$-compression can be achieved by select the first $1-r_{i}$ tokens from the basic solution $f_{cmp}$ generated by $\mathrm{ARC}$:
\begin{equation}
\begin{aligned}
\label{equation:f_comp}
f_{cmp} = \mathrm{ARC}(f_{ori}) \in \mathbb{R}^{D}, 
\end{aligned}
\end{equation}

\begin{equation}
\begin{aligned}
\label{equation:f_comp_ri}
f_{cmp}^{r_i} = f_{cmp}[0:D(1-r_i)] \in \mathbb{R}^{D(1-r_{i})},
\end{aligned}
\end{equation}
where $\mathrm{ARC}(\cdot)$ denotes the arbitrary ratio compressor, which consists of a 12-layer transformer. Fed with the original feature $f_{ori}$, the ARC generates $f_{cmp}$ at the least compression ratio $r_{min}$. Compressed features at other compression ratios $r_i$ can be directly obtained from a prefix of $f_{cmp}$.

\textbf{(2) Feature reconstruction via decoders.}
To ensure robust feature reconstruction across arbitrary compression ratios during training, we introduce a decoder-based reconstruction module. This module is used only during training and discarded at inference, enabling efficient, on-the-fly compression without added computational cost.


Our design centers on a \textit{decoder pool} that contains multiple decoder clusters, each dedicated to a specific compression ratio $r_i$. When the compressed feature $f_{cmp}^{r_i}$ is generated at ratio $r_i$, it is routed to the corresponding decoder cluster. 

Each cluster consists of one main decoder and $M$ auxiliary decoders, as illustrated in Fig. \ref{fig:AuxDecoder}. To generate multi-view representations of the same compressed feature, we apply Monte Carlo \textit{dropout} with different dropout rates $dr_m$, sampled uniformly from $[0.1,0.9]$, producing $M$ stochastic views of $f_{cmp}^{r_i}$. These multi-view features approximate samples from the posterior distribution of the ARC, transforming a deterministic output into a probabilistic one. 
The main decoder reconstructs the primary features and outputs the reconstructed feature $f_{rec}\in \mathbb{R}^{D}$, while each auxiliary decoder reconstructs one of the dropout views, yielding auxiliary outputs $f_{aux-rec}^{m}\in \mathbb{R}^{D}$. All decoders are implemented as a single fully-connected (FC) layer with shape $D(1-r_{i}) \times D$. By assembling multiple decoders to reconstruct these multi-view compressed features, we can effectively reduce the ARC's uncertainty~\cite{gawlikowski2023survey,abdar2021review} and improve the robustness of the compressed features. 
In our implementation, we use $M=5$ auxiliary decoders per cluster. 



The loss functions used for feature reconstruction are the Euclidean distance between the original feature $f_{ori}$ and the reconstructed features $f_{rec}$ and $f_{aux-rec}^{m}$:
\begin{equation}
\begin{aligned}
\label{equation:rec_loss}
\mathcal{L}_{rec} = ||f_{ori}-f_{rec}||_2^2,
\end{aligned}
\end{equation}
\begin{equation}
\begin{aligned}
\label{equation:auxrec_loss}
&\mathcal{L}_{aux-rec} = \sum_{m=1}^M ||f_{ori}-f_{aux-rec}^{m}||_2^2.
\end{aligned}
\end{equation}

\subsection{Mixture of solutions module}\label{sec:MoS}
The Mixture of Solutions (MoS) module enhances the quality of compressed features by refining multiple diverse solutions, \ie, compressed features with different \textit{dropout} views, through a stack of transformer-based MoS blocks (six blocks in our implementation).
Each block performs cross-solution attention, enabling information exchange among solutions and allowing the model to capture complementary patterns. This refinement leads to a more robust and coherent final representation. 
While prior work shows that simple ensemble methods, such as averaging or pooling across multiple outputs, can reduce output uncertainty and improve performance \cite{gawlikowski2023survey,abdar2021review}, MoS goes beyond these methods by learning an optimal way to combine multiple solutions via attention-based refinement.

\textbf{(1) Basic solutions of the MoS.}
To further enhance the representation of compressed features and reduce uncertainty, we use \textit{dropout} to generate multi-view basic solutions from the compressed feature $f_{cmp}$. These solutions are then fed into the Mixture of Solutions (MoS) blocks to obtain a final refined feature. 
In detail, the basic solutions, $\mathbb{S}$, are reshaped as a $K \times D$ feature matrix as follows:
\begin{equation}
\begin{aligned}
\label{equation:solution}
\mathbb{S} = [f_{cmp}^{dr_1 \top}, f_{cmp}^{dr_2 \top}, ... , f_{cmp}^{dr_K \top}], \\ f_{cmp}^{dr_k} \in \mathbb{R}^{D}, \mathbb{S}\in \mathbb{R}^{K \times D},
\end{aligned}
\end{equation}
where $f_{cmp}^{dr_k}$ is the $k$-th solution of the compressed feature $f_{cmp}$, with dropout rate $dr_k$. In our implementation, the dropout rates $dr_k$ for generating multiple views are sampled uniformly from $[0.1,0.9]$. The number of solutions is $K=5$.

\textbf{(2) Cross-solution attention:} 
Each MoS block is built upon a standard transformer block. For the $l$-th block, it receives the basic solutions $\mathbb{S}$, and the output tokens from the $l-1$-th block as the input. Given that $\mathbb{S}\in \mathbb{R}^{K\times D}$ represents $K$ solutions with dimension $D$, it can be regarded as a token sequence of length $K$. The input to the $l$-th block concatenates $\mathbb{S}$ with the previous block's output $\mathrm{MoS}_{l-1}$, forming a sequence of length $2K$. The transformation is formulated as:
\begin{equation}\label{equation:MoS_block}
\begin{aligned}
\mathrm{MoS}_{l} &= \mathrm{FFN}(\mathrm{MHSA}([\mathbb{S}, \mathrm{MoS}_{l-1}]))[K:2K],
\end{aligned}
\end{equation}
where $\mathrm{MHSA}(\cdot)$ is the Multi-Head Self-Attention operation (MHSA) and $\mathrm{FFN}(\cdot)$ is the Feed Forward Net (FFN). And $[K:2K]$ selects the last $K$ tokens as the output of the block. 
Since the number of solutions $K$ is typically small (\eg, less than 10), the token sequence length in MoS remains much shorter than in commonly used Transformers, resulting in negligible computational overhead and minimal impact on overall model efficiency.

\textbf{(3) Compression token:} 
Inspired by the classification token in transformers, a learnable compression token $\mathbb{C}\in \mathbb{R}^D$ is introduced to aggregate knowledge across all solutions. During training, $\mathbb{C}$ is inserted into the input of each MoS block, increasing the input sequence length from $2K$ to $2K+1$. This allows the compression token to attend to all solution tokens and gradually accumulate the most informative features through the layers. 
After passing through $L$ MoS blocks, the final representation of the compression token is taken as the final compressed feature:
\begin{equation}\label{equation:final_solution}
f_{cmp*} = \mathbb{C}^{(L)},
\end{equation}
where $\mathbb{C}^{(L)}$ denotes the compression token of the $L$-th MoS block. 
Furthermore, the final compressed feature $f_{cmp*}$ is recovered by a main decoder and $M$ auxiliary decoders. These decoders are implemented as a single FC layer and are trained using loss functions defined in Eqn. (\ref{equation:rec_loss}) and Eqn. (\ref{equation:auxrec_loss}).

\subsection{Entity relation graph constraint} \label{sec:ERGC}
To preserve the geometric structure and relationships within the original feature space during compression, we introduce the Entity Relation Graph Constraint (ERGC). While reconstruction loss focuses on recovering the original feature from its compressed version, it does not inherently maintain the structural relationships among entities. 
ERGC addresses this by ensuring that these relationships are preserved in the compressed feature space, thereby enhancing the discriminative power and representation capability of the compressed features. 

\begin{figure}
  \centering
  \includegraphics[width=0.92\linewidth]{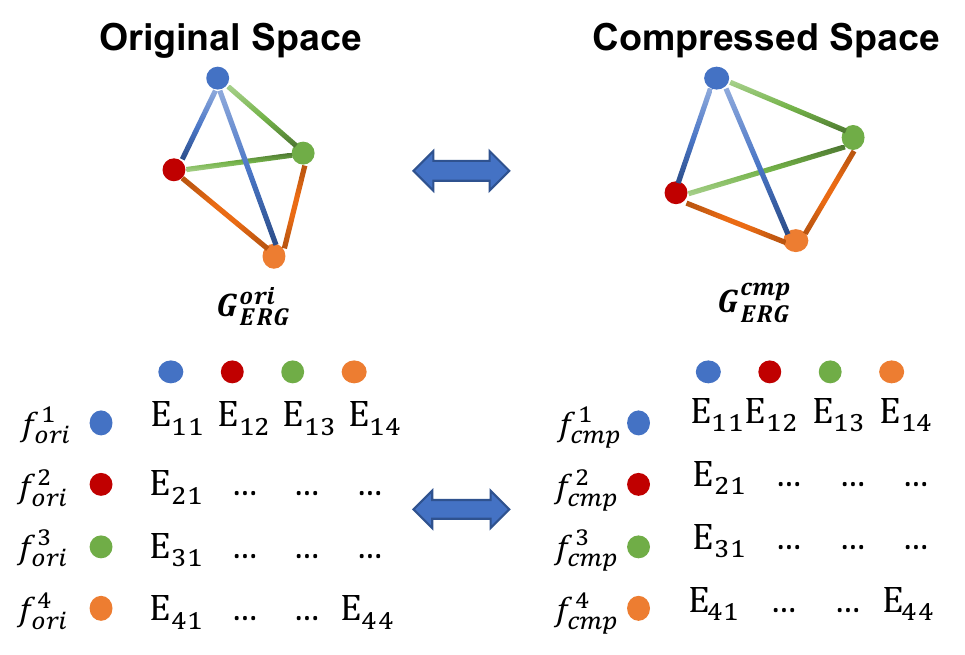}
  \caption{An illustration of the Entity Relation Graph Constraint (ERGC). It constructs two Entity-Relation Graphs (ERG) for the original feature space and the compressed feature space, respectively. Then, it forces the latter to be as close as the former. Note that ``$E_{12}$'' represents the relationship between the first entity and the second entity.}
  \label{fig:ERGC} 
\end{figure}

\textbf{(1) ERG construction:} 
As illustrated in Fig. \ref{fig:ERGC}, the ERGC module constructs two Entity-Relation Graphs (ERG): one for the original feature space ($G^{ori}_{ERG}=(\mathcal{V}_{ori}, \mathcal{E}_{ori})=(\mathbf{f}_{ori}, \mathbb{E}_{ori})$) and another for the compressed feature space ($G^{cmp}_{ERG}=(\mathcal{V}_{cmp}, \mathcal{E}_{cmp})=(\mathbf{f}_{cmp}, \mathbb{E}_{cmp})$). Note that $\mathcal{V}$ denotes the vertex set of the graph and $\mathcal{E}$ is the edge set of the graph. 
Each vertex in these graphs represents the feature $\mathbf{f}$ of an entity (\eg, images or texts), and each edge value is calculated using the cosine similarity between pairs of entities. Specifically, for any pair of entities $a$ and $b$, their relationship (\ie, the edge between vertices $a$ and $b$) is assigned the value:
\begin{equation}
\begin{aligned}
\label{equation:erg}
\mathbb{E}(a, b) = <f_a, f_b>,
\end{aligned}
\end{equation}
in which $f_{a}$ and $f_{b}$ represent the feature of entity $a$ and entity $b$ respectively, and $<\cdot,\cdot>$ is the cosine similarity. For a batch of $B$ samples, the edges of ERG can be formulated as a $B \times B$ matrix, each element of which is the cosine similarity between two entities. 

\textbf{(2) ERGC imposing:}  
In both the ARC and MoS modules, the training process imposes constraints on the vertices of the ERG (\ie, the feature reconstruction constraints), ensuring that individual entities are accurately reconstructed. However, ERGC goes further by constraining the edge values of the graph, which represent the relationships between entities. Specifically, ERGC minimizes the difference between the edges of the original and compressed feature spaces:
\begin{equation}
\mathcal{L}_\text{ERGC} = ||\mathbb{E}_{ori} - \mathbb{E}_{cmp}||_2^2,
\label{eqn:ERGC_loss}
\end{equation}
where $\mathbb{E}_{ori}$ and $\mathbb{E}_{cmp}$ are the adjacency matrices containing the edge weights (cosine similarities) of the original and compressed ERGs, respectively.

\subsection{Optimization}\label{sec:optimization}
This subsection describes the overall optimization strategy of the proposed framework. The training is conducted first for the ARC and then for the MoS. 

\textbf{(1) ARC optimization:} 
We train the ARC module to learn one compressed feature with multiple compression ratios $r_i$. For each ratio, the compressed features $f_{cmp}^{r_i}$ are routed to a decoder cluster, which consists of a main decoder and several auxiliary decoders to reconstruct the original feature. The reconstruction losses in Eqn. (\ref{equation:rec_loss}) and Eqn. (\ref{equation:auxrec_loss}) are imposed. We also apply the ERGC loss (Eqn. (\ref{eqn:ERGC_loss})) to preserve semantic and geometric relationships among entities. 
Thus, the total loss for ARC optimization is formulated as:
\begin{equation}
\begin{aligned}
\label{equation:ARC_obj}
\mathcal{L}_\text{ARC} = \sum_{r_i} \left( \right.&\mathcal{L}_{rec}(f_{cmp}^{r_i}) + \frac{1}{M}\mathcal{L}_{aux-rec}(f_{cmp}^{r_i}) \\
& \left. + \lambda \mathcal{L}_\text{ERGC}(f_{cmp}^{r_i}) \right),
\end{aligned}
\end{equation}
where $M$ is the number of auxiliary decoders and $\lambda$ is the penalty weight of $L_\text{ERGC}$. And $\mathcal{L}_{rec}(f_{cmp}^{r_i})$, $\mathcal{L}_{aux-rec}(f_{cmp}^{r_i})$, and $\mathcal{L}_\text{ERGC}(f_{cmp}^{r_i})$ denote the reconstruction loss, auxiliary reconstruction loss and ERGC loss for the compressed feature $f_{cmp}^{r_i}$, respectively.

\textbf{Progressive compression training.} 
To enable flexible and stable learning across compression ratios, we present the Progressive Compression Training Scheme, which adopts a dynamic sampling strategy during training. At each step, we randomly sample a compression ratio $r_i \in [0,1]$ using the dynamic sampling strategy, and reconstruct the corresponding truncated compressed feature $f_{cmp}^{r_i}$ (Eqn. (\ref{equation:f_comp_ri})) using both the reconstruction loss and the ERGC loss (Eqn. (\ref{equation:ARC_obj})). This ensures that the model learns to compress features effectively at any target ratio. 

Since ARC generates compressed features in a token-by-token auto-regressive manner, early output tokens correspond to high compression ratios (\ie, fewer retained tokens). Therefore, training begins by focusing on high compression ratios to ensure the model first learns to preserve the most critical semantic information. As training progresses, the sampling distribution gradually shifts to favor lower compression ratios, enabling the model to refine and enrich the compressed representation with additional tokens.
The sampling follows a beta distribution:
\begin{equation}
\begin{aligned}
\label{equation:beta_distribution}
p(r_i;\alpha,\beta) = \frac{r_i^{\alpha-1}(1-r_i)^{\beta-1}}{B(\alpha,\beta)}, \\
\text{for} \quad r_i\in [0,1],
\end{aligned}
\end{equation}
where $B(\alpha,\beta)$ is the beta function for normalization. The parameters $\alpha$ and $\beta$ control the concentration of the distribution near 1 and 0, respectively. Initially, we set $\alpha=80, \beta=5$, resulting in a highly right-skewed distribution that favors high compression ratios. As training proceeds, we gradually decrease $\alpha$, shifting the distribution toward lower compression ratios. Eventually, 
$\alpha$ and $\beta$ are set to equal values, resulting in a more uniform distribution. This ensures that in later stages, a wide range of compression ratios are adequately sampled, enabling robust generalization across all target ratios at inference.

\textbf{(2) MoS optimization:} 
We refine the compressed features using the MoS module. In this stage, we first generate the basic compressed feature $f_{cmp}\in\mathbb{R}^D$ by using the full input dimension (\ie, compression ratio $r_{min}=0$). This feature is then refined through cross-solution attention and aggregated via a compression token to produce the final output feature $f_{cmp*}\in\mathbb{R}^D$. 
Similar to ARC optimization, we reconstruct the compressed features at different compression ratios $f_{cmp*}^{r_i}$ using the reconstruction loss and the ERGC loss. The total loss for MoS optimization is formulated as:
\begin{equation}
\begin{aligned}
\label{equation:MoS_obj}
\mathcal{L}_\text{MoS} = 
\sum_{r_i} \left( \right.&\mathcal{L}_{rec}(f_{cmp*}^{r_i}) + \frac{1}{M}\mathcal{L}_{aux-rec}(f_{cmp*}^{r_i}) \\
& \left. + \lambda \mathcal{L}_\text{ERGC}(f_{cmp*}^{r_i}) \right).
\end{aligned}
\end{equation}
Since the compressed representation already supports arbitrary compression ratios after ARC training, we do not use progressive sampling during MoS optimization. Instead, we sample compression ratios uniformly from [0,1], corresponding to setting $\alpha=\beta=1$ in the beta distribution.

\subsection{Training procedure}\label{sec:procedure}

The overall training procedure of the proposed Arbitrary Ratio Feature Compression (ARFC) framework is summarized in Alg.~\ref{alg:ARFC}. The method consists of two main stages: ARC training and MoS training, both based on reconstruction and ERGC losses.

In the first stage, we extract original features $f_{\text{ori}}$ using a frozen feature encoder. Then, the ARC module is trained using reconstruction and ERGC losses, to generate compressed features $f_{\text{cmp}} \in \mathbb{R}^D$ that support arbitrary compression ratios. During this phase, we employ a \textit{progressive compression training scheme}: at each step, a compression ratio $r_i \in [0,1]$ is sampled from a Beta distribution with parameters $\alpha$ and $\beta$. These parameters are gradually adjusted during training, starting from $\alpha >\beta$ (favoring high compression) and progressively decreasing $\beta$ until $\alpha = \beta$, resulting in a more uniform sampling of compression ratios. This schedule enables coarse, high-compression representations to finer, low-compression ones, ensuring robustness across all target ratios.

Once the ARC is trained, we proceed to optimize the MoS module. Similar to ARC, it uses reconstruction and ERGC losses for refinement. However, since the compressed representation already supports arbitrary compression ratios, we sample $r_i$ uniformly from $[0,1]$ (\ie, $ \alpha = \beta = 1 $) during MoS training, ensuring robustness across all compression levels without progressive adaptation.

The final output of the full network (ARC followed by MoS) is the refined compressed feature $f_{\text{cmp}^*}$, which maintains high quality under any desired compression ratio.

\begin{algorithm}[!htbp]
	\caption{The arbitrary ratio feature compression algorithm.}
	\label{alg:ARFC}
    \LinesNumbered
		\KwIn{Dataset $\mathcal{D}^\mathrm{train}=\{\mathbf{x},\mathbf{y}\}$, ARC model $\mathbf{\Theta}^\mathrm{ARC}$, MoS model $\mathbf{\Theta}^\mathrm{MoS}$, Decoder pool $\{\mathbf{\Theta}^\mathrm{D}_{r_i}\}$ indexed by compression ratio $r_i$.}
        Extract original features $\mathbf{f}_{ori}$ of the dataset $\mathcal{D}^\mathrm{train}$ using a frozen feature encoder.\\
		\Repeat{ARC training converges}{
            Generate compressed features $f_{cmp}$ via ARC fed with the original feature $f_{ori}$ (Eqn. (\ref{equation:f_comp}));\\
            Sample compression ratios $\{r_i\}$ via \textit{progressive compression training scheme};\\
            \For {each sampled compression ratio $r_i$}{
                Compute $f_{cmp}^{r_i}$ using Eqn. (\ref{equation:f_comp_ri});\\
                Route a corresponding decoder cluster $\mathbf{\Theta}^\mathrm{D}_{r_i}$ given $r_i$;\\
                Compute the corresponding reconstructed features $f_{rec}$ and $\{f_{aux\text{-}rec}^m\}_{m=1}^M$ from the main decoder and the auxiliary decoders;\\
                Update $\mathbf{\Theta}^\mathrm{ARC}$ and $\{\mathbf{\Theta}^\mathrm{D}_{r_i}\}$ using Eqn. (\ref{equation:ARC_obj});
            } 
        }
        \Repeat{MoS training converges}{
            Compute the final solution $f_{cmp*}$ of the compressed feature via MoS using Eqns. (\ref{equation:MoS_block}) and (\ref{equation:final_solution});\\
            \For{each sampled compression ratio $r_i$}{
                Compute $f_{cmp*}^{r_i}$ using Eqn. (\ref{equation:f_comp_ri});\\
                Update $\mathbf{\Theta}^\mathrm{MoS}$ using Eqn. (\ref{equation:MoS_obj});
            }
		}
		\Return $\mathbf{\Theta}^\mathrm{ARC}$ and $\mathbf{\Theta}^\mathrm{MoS}$.		
\end{algorithm}

\section{Experiments}\label{experiments}
In this section, we conduct comprehensive experiments to validate the effectiveness of the proposed method. Sec. \ref{sec:settings} describes the experimental settings and implementation details. Sec. \ref{sec:evaluation} evaluates the proposed method on various tasks. Additionally, Sec. \ref{sec:ablation} provides an ablation study to analyze deeply each component in the proposed framework.

\subsection{Settings}\label{sec:settings}

\textbf{Datasets and Tasks.}
We evaluate the proposed method on three types of tasks: cross-modal retrieval \cite{wang2016comprehensive}, image classification \cite{rawat2017deep}, and image retrieval \cite{dubey2021decade}. For cross-modal retrieval, we use the Flickr30K-CN \cite{lan2017fluency} for Chinese and Flickr30K \cite{plummer2015flickr30k} for English datasets. 
For image classification, we evaluate on the following datasets: CIFAR 10/100 \cite{krizhevsky2009learning}, ImageNet \cite{russakovsky2015imagenet},  DTD \cite{cimpoi2014describing}, EuroSAT \cite{helber2019eurosat}, FER \cite{goodfellow2013challenges}, FGVC \cite{maji2013fine}, KITTI \cite{geiger2013vision}, MNIST \cite{deng2012mnist}, PC \cite{veeling2018rotation}, and VOC \cite{everingham2010pascal}. 
For image retrieval, we use the CUB-200-2011~\cite{wah2011caltech} and Cars-196~\cite{KrauseStarkDengFei-Fei_3DRR2013} datasets. 
In all experiments, a series of CLIP models \cite{radford2021learning} are used as baselines, including CN-CLIP \cite{yang2022chinese} and SigLip 2 \cite{tschannen2025siglip}. The ViT-H/14 and ViT-B/16 architectures are used as the backbones. The original features extracted from these baseline models serve as performance baselines.

\textbf{Compared Methods.}
We compare our method with several state-of-the-art feature compression methods from different paradigms: Q-Former \cite{li2023blip}, Autoencoder \cite{pinheiro2021variational}, and Post-Training Quantization (PTQ) \cite{liu2021post} techniques. 
For \textit{Q-Former}, we adopt the strategy in BLIP2 \cite{li2023blip}, which compresses features by reducing the sequence length through a learnable query mechanism.
For \textit{Autoencoders}, we use the Variational Autoencoder (VAE) \cite{pinheiro2021variational}, a widely used approach that learns to encode input features into a lower-dimensional latent space and then reconstructs them via a decoder.
For \textit{PTQ} methods, we compare MinMax quantization \cite{schuster1999review} and P4Q \cite{sun2024p4q}, which convert pre-trained models into lower-precision representations without requiring retraining.
All compared methods are evaluated under the same compression ratios and backbone architecture as our method, ensuring a fair and consistent comparison.

\textbf{Implementation Details.}
During training, we use a batch size of 1024 (per GPU). The learning rate is initialized to 0.001 and is dynamically controlled by an AdamW optimizer~\cite{loshchilov2017decoupled} with the default settings of weight decay and the beta parameters. 
The hyperparameter $\lambda$ in Eqns. (\ref{equation:ARC_obj}) and (\ref{equation:MoS_obj}) is selected by grid search, which yields $\lambda=0.5$.
All experiments are conducted on a platform with 16 Nvidia H200 GPU cards. Training is conducted using PyTorch 2.6.0~\cite{paszke2019pytorch}, leveraging the Hugging Face Transformers library~\cite{wolf2020transformers} for model implementations.


\subsection{Evaluation Results}\label{sec:evaluation}
We evaluate the proposed method and compare it with state-of-the-art methods on cross-modal retrieval, image classification, and image retrieval. 
The original feature has a dimension of 1024. We reduce the feature dimension to $1/2$, $1/4$, $1/8$, and $1/16$, which correspond to the compression ratios of 50\%, 75\%, 87.5\%, and 93.75\%, respectively.

\textbf{(1) Evaluation on Cross-modal Retrieval.} 
Tab. \ref{tab:crossRetrieval_compare} reports the evaluation (measured by Recall at 1 (R@1) and Recall at 5 (R@5)) on the cross-modal retrieval task. The experiments are conducted on both the English and Chinese versions of the Flickr30K dataset. Results for both zero-shot and fine-tuned scenarios are included to provide a comprehensive view of our method's effectiveness. From this table, our method consistently outperforms compared methods, including autoencoder-based approaches, Q-former, and PTQ, across various compression ratios. 
Our method achieves competitive results under zero-shot conditions. With fine-tuning, the performance further improves, demonstrating the potential for customization to specific tasks or datasets. Remarkably, at a 50\% compression ratio, our method even surpasses the R@1 score of uncompressed features (\ie, Baseline), indicating enhanced retrieval accuracy despite a reduced dimension. 
Whether applied to the English or Chinese version of Flickr30K and regardless of the backbone architecture used, our method shows significant improvements. This consistency highlights the adaptability and broad applicability of our framework in diverse settings. 

\begin{table*}[!htbp]
	\small
	\centering
	\setlength\tabcolsep{8pt}
	\setlength{\extrarowheight}{5pt}
	\caption{Evaluation on cross-modal retrieval. On Flickr30K, the backbone model for feature extraction is SigLip 2 with ViT-B/16 architecture. On Flickr30K-CN, the backbone model is CN-CLIP, utilizing the ViT-H/14 architecture. The ``Baseline'' in the table represents the original feature whose dimension is 1024. Since PTQ requires training data for accuracy alignment during quantization, we include its results in the ``fine-tuning'' comparison with other methods. The blue-marked results indicate performance that matches or surpasses the baseline.}
	\resizebox{1\textwidth}{!}{
		\begin{tabular}{|c|c|ccc|ccc|ccc|ccc|} 
			\hline
            \rowcolor{mygray}\multicolumn{14}{|c|}{\textbf{Dataset: \quad Flickr30K}}  \\ \hline
            \multicolumn{2}{|c|}{\textbf{Task}} & \multicolumn{6}{|c|}{\textbf{Text-to-Image}} & \multicolumn{6}{|c|}{\textbf{Image-to-Text}}  \\ \hline
            \multicolumn{2}{|c|}{\textbf{Setting}} & \multicolumn{3}{|c|}{\textbf{Zero-Shot}} & \multicolumn{3}{|c|}{\textbf{Finetuning}} & \multicolumn{3}{|c|}{\textbf{Zero-Shot}} & \multicolumn{3}{|c|}{\textbf{Finetuning}}   \\ \hline
            Method & CR & R@1 & R@5 & R@10 & R@1 & R@5 & R@10 & R@1 & R@5 & R@10 & R@1 & R@5 & R@10   \\ \hline
            Baseline [SigLip 2 (ViT-B/16)] & 0 & 84.5 & 95.8 & 98.0 & 89.6 & 98.8 & 99.7 & 94.4 & 99.5 & 100 & 98.7 & 100 & 100 \\ \hline
            AutoEncoder-VAE & 50\% & 83.3 & 95.1 & 97.6 & 88.9 & 98.5 & 99.5 & 94.1 & 99.3 & 100 & 98.5 & 100 & 100 \\
            Q-Former & 50\% & 82.1 & 94.6 & 97.3 & 88.5 & 98.3 & 99.5 & 93.2 & 99.0 & 100 & 98.1 & 100 & 100 \\
            \textbf{Ours} & 50\% & \textbf{\OPBL{84.8}} & \textbf{\OPBL{96.0}} & \textbf{\OPBL{98.1}} & \textbf{\OPBL{89.7}} & \textbf{\OPBL{98.8}} & \textbf{\OPBL{99.8}} & \textbf{\OPBL{94.6}} & \textbf{\OPBL{99.5}} & \textbf{\OPBL{100}} & \textbf{\OPBL{98.9}} & \textbf{\OPBL{100}} & \textbf{\OPBL{100}} \\ \hline
            AutoEncoder-VAE & 75\% & 81.7 & 94.4 & 97.3 & 87.8 & 98.1 & 99.3 & 91.1 & 98.5 & 99.7 & 97.5 & 99.8 & 100 \\
            Q-Former & 75\% & 81.2 & 94.3 & 97.1 & 87.1 & 97.8 & 99.3 & 91.6 & 98.7 & 99.7 & 97.8 & 99.9 & 100 \\
            PTQ-MinMax (8bit) & 75\% & - & - & - & 78.3 & 94.2 & 97.4 & - & - & - & 88.2 & 98.3 & 99.3 \\
            PTQ-P4Q (8bit) & 75\% & - & - & - & 78.8 & 94.4 & 97.5 & - & - & - & 88.8 & 98.4 & 99.3 \\
            \textbf{Ours} & 75\% & \textbf{84.3} & \textbf{\OPBL{95.8}} & \textbf{97.9} & \textbf{89.5} & \textbf{\OPBL{98.8}} & \textbf{99.6} & \textbf{\OPBL{94.4}} & \textbf{\OPBL{99.5}} & \textbf{\OPBL{100}} & \textbf{98.8} & \textbf{\OPBL{100}} & \textbf{\OPBL{100}} \\ \hline
            AutoEncoder-VAE & 87.5\% & 72.3 & 91.8 & 95.6 & 82.9 & 96.6 & 98.4 & 81.8 & 97.4 & 98.8 & 95.6 & 99.6 & 100  \\
            Q-Former & 87.5\% & 71.3 & 91.4 & 95.4 & 79.9 & 93.8 & 97.6 & 81.5 & 97.3 & 98.7 & 95.3 & 99.4 & 100 \\
            PTQ-MinMax (4bit) & 87.5\% & - & - & - & 72.1 & 91.7 & 95.2 & - & - & - & 82.3 & 96.5 & 98.3 \\
            PTQ-P4Q (4bit) & 87.5\% & - & - & - & 73.5 & 92.3 & 95.6 & - & - & - & 84.1 & 97.4 & 98.7 \\
            \textbf{Ours} & 87.5\% & \textbf{80.3} & \textbf{94.1} & \textbf{97.0} & \textbf{88.8} & \textbf{98.6} & \textbf{99.6} & \textbf{92.5} & \textbf{98.9} & \textbf{99.9} & \textbf{97.7} & \textbf{99.9} & \textbf{\OPBL{100}}\\ \hline
            AutoEncoder-VAE & 93.75\% & 62.7 & 86.8 & 92.7 & 72.5 & 91.5 & 95.4 & 75.6 & 94.4 & 97.0 & 93.3 & 99.0 & 99.5 \\
            Q-Former & 93.75\% & 60.3 & 86.1 & 92 & 71.8 & 91.2 & 95.3 & 74.2 & 94.1 & 96.8 & 93.1 & 99.1 & 99.4 \\
            \textbf{Ours} & 93.75\% & \textbf{71.6} & \textbf{91.1} & \textbf{95.1} & \textbf{78.7} & \textbf{93.2} & \textbf{97.1} & \textbf{87.3} & \textbf{98.1} & \textbf{99.6} & \textbf{95.3} & \textbf{99.5} & \textbf{\OPBL{100}}\\ \hline
            \multicolumn{14}{|c|}{\textbf{Dataset: \quad Flickr30K-CN}}  \\ \hline
            \multicolumn{2}{|c|}{\textbf{Task}} & \multicolumn{6}{|c|}{\textbf{Text-to-Image}} & \multicolumn{6}{|c|}{\textbf{Image-to-Text}}  \\ \hline
            \multicolumn{2}{|c|}{\textbf{Setting}} & \multicolumn{3}{|c|}{\textbf{Zero-Shot}} & \multicolumn{3}{|c|}{\textbf{Finetuning}} & \multicolumn{3}{|c|}{\textbf{Zero-Shot}} & \multicolumn{3}{|c|}{\textbf{Finetuning}}   \\ \hline
            Method & CR & R@1 & R@5 & R@10 & R@1 & R@5 & R@10 & R@1 & R@5 & R@10 & R@1 & R@5 & R@10   \\ \hline
            Baseline [CN-CLIP (ViT-H/14)] & 0 & 71.2 & 91.4 & 95.5 & 83.8 & 96.9 & 98.6 & 81.6 & 97.5 & 98.8 & 95.3 & 99.7 & 100 \\
			\hline
            AutoEncoder-VAE & 50\% & 69.8 & 90.6 & 95.0 & 83.1 & 96.7 & 98.5 & 81.0 & 97.3 & 98.7 & 95.0 & 99.3 & 100 \\
            Q-Former & 50\% & 68.1 & 89.4 & 94.7 & 82.3 & 95.9 & 98.1 & 79.1 & 96.7 & 98.4 & 93.6 & 99.1 & 99.9 \\
            \textbf{Ours} & 50\% & \textbf{71.1} & \textbf{\OPBL{91.4}} & \textbf{\OPBL{95.6}} & \textbf{\OPBL{84.3}} & \textbf{\OPBL{97.1}} & \textbf{\OPBL{98.7}} & \textbf{\OPBL{81.7}} & \textbf{\OPBL{97.5}} & \textbf{\OPBL{98.8}} & \textbf{\OPBL{95.5}} & \textbf{\OPBL{99.8}} & \textbf{\OPBL{100}} \\ \hline
            AutoEncoder-VAE & 75\% & 68.3 & 89.5 & 94.7 & 81.7 & 95.7 & 98 & 79.5 & 96.8 & 98.5 & 93.8 & 99.2 & 100 \\
            Q-Former & 75\% & 65.7 & 88.1 & 93.9 & 80.3 & 95.4 & 97.9 & 75.8 & 94.1 & 97.9 & 92.7 & 98.9 & 99.7 \\
            PTQ-MinMax (8bit) & 75\% & - & - & - & 72.8 & 91.5 & 95.5 & - & - & - & 87.3 & 97.9 & 99.2 \\
            PTQ-P4Q (8bit) & 75\% & - & - & - & 73.1 & 91.6 & 95.6 & - & - & - & 87.8 & 98.1 & 99.2 \\
            \textbf{Ours} & 75\% & \textbf{70.2} & \textbf{90.8} & \textbf{95.2} & \textbf{83.4} & \textbf{\OPBL{96.9}} & \textbf{98.5} & \textbf{80.8} & \textbf{97.3} & \textbf{98.6} & \textbf{\OPBL{95.3}} & \textbf{99.4} & \textbf{\OPBL{100}} \\ \hline
            AutoEncoder-VAE & 87.5\% & 63.8 & 87.2 & 93.1 & 79.2 & 94.8 & 97.3 & 76.1 & 94.2 & 97.9 & 93.1 & \textbf{99} & 99.6 \\
            Q-Former & 87.5\% & 61.9 & 86.5 & 92.3 & 77.5 & 93.8 & 96.9 & 74.9 & 93.6 & 97.3 & 90.2 & 98.8 & 99.5 \\
            PTQ-MinMax (4bit) & 87.5\% & - & - & - & 65.8 & 88.4 & 93.2 & - & - & - & 81.4 & 96.2 & 98.5 \\
            PTQ-P4Q (4bit) & 87.5\% & - & - & - & 66.4 & 88.8 & 93.4 & - & - & - & 81.7 & 86.3 & 98.5 \\
            \textbf{Ours} & 87.5\% & \textbf{68.1} & \textbf{89.4} & \textbf{94.5} & \textbf{81.8} & \textbf{95.8} & \textbf{97.9} & \textbf{79.5} & \textbf{96.7} & \textbf{98.3} & \textbf{93.6} & 98.9 & \textbf{99.9}\\ \hline
            AutoEncoder-VAE & 93.75\% & 52.5 & 79.1 & 86.9 & 67.8 & 89.4 & 93.7 & 63.7 & 86.8 & 93.1 & 82.9 & 97.1 & 98.9 \\
            Q-Former & 93.75\% & 49.7 & 77.2 & 84.8 & 66.4 & 88.6 & 93.5 & 60.5 & 85.8 & 92.3 & 83.1 & 96.8 & 98.8 \\
            \textbf{Ours} & 93.75\% & \textbf{60.5} & \textbf{85.2} & \textbf{91.4} & \textbf{73.4} & \textbf{92.1} & \textbf{96.0} & \textbf{74.7} & \textbf{93.7} & \textbf{97.2} & \textbf{89.9} & \textbf{98.7} & \textbf{99.6} \\ \hline
	\end{tabular}  }
	\label{tab:crossRetrieval_compare}
\end{table*}

\textbf{(2) Evaluation on Image Classification.}
We further evaluate the quality of compressed features on image classification tasks across 11 diverse datasets, including both general and fine-grained recognition benchmarks. The evaluation is conducted in a zero-shot setting, where the compressed features are directly evaluated without fine-tuning, providing a clear measure of how well the semantic information is preserved.

As shown in Tab. \ref{tab:classification_compare}, our method achieves superior performance compared with autoencoder-based models, Q-former, and PTQ, across various compression ratios. P4Q achieves good performance because it uses training data for calibration, giving it an advantage over other zero-shot methods.
Notably, at a 50\% compression ratio, our method achieves better accuracy than the original uncompressed features on most datasets. Even under a high compression ratio of 75\%, our approach still surpasses the performance of the full-dimensional features on ImageNet, demonstrating its strong representational capability. Moreover, the effectiveness of our method is consistently observed across different backbone architectures and different datasets. This indicates that our framework is robust to model variations and data domains. 

\begin{table*}[!htbp]
	\small
	\centering
	\setlength\tabcolsep{8pt}
	\setlength{\extrarowheight}{5pt}
	\caption{Evaluation on image classification. The ``Baseline'' in the table represents the original feature whose dimension is 1024. The ``FT'' denotes ``Fine-tuning'', which indicates that PTQ methods require training data for accuracy alignment. 
    The blue-marked results indicate performance that matches or surpasses the baseline.}
	\resizebox{1\textwidth}{!}{
		\begin{tabular}{|c|c|ccccccccccc|} 
			\hline
            \rowcolor{mygray}\multicolumn{13}{|c|}{\textbf{Baseline model: \quad SigLip 2 (ViT-B/16)}}  \\ \hline
            \multicolumn{2}{|c|}{\textbf{Dataset}} & CIFAR10  & CIFAR100 & ImageNet & DTD & EuroSAT & FER & FGVC & KITTI & MNIST & PC & VOC  \\ \hline
            Method & CR & \multicolumn{11}{|c|}{\textbf{Test accuracy (Zero-shot)}} \\ \hline
            Baseline [SigLip 2 (ViT-B/16)] & 0 & 98.4\% & 83.1\% & 81.2\% & 59.4\% & 56.4\% & 56.5\% & 28.8\% & 49.9\% & 81.7\% & 64.4\% & 85.7\% \\ \hline
            AutoEncoder-VAE & 50\% & 97.9\% & 82.7\% & 80.5\% & 59\% & 55.9\% & 56.3\% & 28.7\% & 49.6\% & 81.6\% & 63.8\% & 84.9\% \\
            Q-Former & 50\% & 97.5\% & 81.3\% & 80.1\% & 58.5\% & 54.8\% & 55.8\% & 28.2\% & 49.2\% & 81.2\% & 63.4\% & 84.7\% \\
            \textbf{Ours} & 50\% & \textbf{\OPBL{98.5\%}} & \textbf{\OPBL{83.3\%}} & \textbf{\OPBL{81.7\%}} & \textbf{59.3\%} & \textbf{\OPBL{56.7\%}} & \textbf{\OPBL{56.8\%}} & \textbf{\OPBL{29\%}} & \textbf{\OPBL{50.1\%}} & \textbf{\OPBL{81.8\%}} & \textbf{\OPBL{64.4\%}} & \textbf{\OPBL{85.7\%}} \\ \hline
            AutoEncoder-VAE & 75\% & 96.2\% & 73.9\% & 73.9\% & 56.8\% & 53.5\% & 55.3\% & 28.3\% & 48.8\% & 80.7\% & 62.3\% & 83.9\% \\ 
            Q-Former & 75\% & 95.9\% & 73.6\% & 74.1\% & 55.7\% & 52.9\% & 54.8\% & 27.9\% & 47.5\% & 80.2\% & 62.5\% & 83.9\% \\ 
            PTQ-MinMax (8bit)-FT & 75\% & 96.4\% & 79.2\% & 78.6\% & 57.3\% & 54.7\% & 54.9\% & 27.7\% & 49.8\% & 80.1\% & 62.6\% & 84.8\% \\ 
            PTQ-P4Q (8bit)-FT & 75\% & 96.9\% & 79.9\% & 80.2\% & 58.1\% & 54.9\% & 55.5\% & 27.9\% & \textbf{\OPBL{50.3\%}} & 80.8\% & 63.5\% & 85.2\% \\
            \textbf{Ours} & 75\% & \textbf{98.3\%} & \textbf{82.9\%} & \textbf{\OPBL{81.5\%}} & \textbf{58.9\%} & \textbf{\OPBL{56.5\%}} & \textbf{\OPBL{56.6\%}} & \textbf{28.6\%} & {49.7\%} & \textbf{81.3\%} & \textbf{64.0\%} & \textbf{85.4\%} \\ \hline
            AutoEncoder-VAE & 87.5\% & 94.9\% & 72.5\% & 72.3\% & 53.4\% & 50.7\% & 52.7\% & 25.7\% & 46.7\% & 77.9\% & 61.4\% & 82.7\% \\
            Q-Former & 87.5\% & 94.5\% & 72.8\% & 72.7\% & 52.1\% & 49.8\% & 52.2\% & 25.1\% & 46.3\% & 78.0\% & 61.6\% & 82.9\% \\
            PTQ-MinMax (4bit)-FT & 87.5\% & 88.5\% & 72.7\% & 69.1\% & 45.8\% & 42.4\% & 44.3\% & 19.1\% & 41.7\% & 69.8\% & 57.4\% & 77.3\% \\
            PTQ-P4Q (4bit)-FT & 87.5\% & 89.7\% & 74.2\% & 70.5\% & 46.3\% & 44.1\% & 44.8\% & 19.7\% & 42.3\% & 71.2\% & 57.7\% & 77.9\% \\
            \textbf{Ours} & 87.5\% & \textbf{97.6\%} & \textbf{80.3\%} & \textbf{80.3\%} & \textbf{55.4\%} & \textbf{55.6\%} & \textbf{54.5\%} & \textbf{27.4\%} & \textbf{48.5\%} & \textbf{80.5\%} & \textbf{62.1\%} & \textbf{84.6\%} \\ \hline
            AutoEncoder-VAE & 93.75\% & 92.2\% & 70.3\% & 70.5\% & 48.3\% & 45.7\% & 48.7\% & 20.4\% & 41.3\% & 73.1\% & 57.4\% & 77.9\% \\
            Q-Former & 93.75\% & 91.4\% & 70.5\% & 68.9\% & 47.6\% & 44.1\% & 47.6\% & 21.8\% & 42.1\% & 72.8\% & 56.1\% & 76.1\% \\
            \textbf{Ours} & 93.75\% & \textbf{95.8\%} & \textbf{73.5\%} & \textbf{74.7\%} & \textbf{51.2\%} & \textbf{50.8\%} & \textbf{51.7\%} & \textbf{23.9\%} & \textbf{44.1\%} & \textbf{73.9\%} & \textbf{59.7\%} & \textbf{79.3\%} \\ \hline
            \multicolumn{13}{|c|}{\textbf{Baseline model: \quad CN-CLIP (ViT-H/14)}}  \\ \hline
            \multicolumn{2}{|c|}{\textbf{Dataset}} & CIFAR10  & CIFAR100 & ImageNet & DTD & EuroSAT & FER & FGVC & KITTI & MNIST & PC & VOC  \\ \hline
            Method & CR & \multicolumn{11}{|c|}{\textbf{Test accuracy (Zero-shot)}} \\ \hline
            Baseline [CN-CLIP (ViT-H/14)] & 0 & 96.0\% & 79.7\% & 59.6\% & 51.2\% & 52.0\% & 55.1\% & 26.2\% & 49.9\% & 79.4\% & 63.5\% & 84.9\%  \\
			\hline
            AutoEncoder-VAE & 50\% & 95.8\% & 79.3\% & 59.4\% & 50.7\% & 51.1\% & 53.8\% & 25.7\% & 49.6\% & 79.3\% & 63.1\% & 84.5\% \\
            Q-Former & 50\% & 95.3\% & 78.2\% & 58.4\% & 49.8\% & 49.9\% & 53.6\% & 24.9\% & 49.2\% & 78.6\% & 62.5\% & 84.4\% \\
            \textbf{Ours} & 50\% & \textbf{\OPBL{96.3\%}} & \textbf{\OPBL{79.7\%}} & \textbf{\OPBL{60.1\%}} & \textbf{51.1\%} & \textbf{51.8\%} & \textbf{55.1\%} & \textbf{\OPBL{26.4\%}} & \textbf{\OPBL{50.0\%}} & \textbf{\OPBL{79.4\%}} & \textbf{63.4\%} & \textbf{\OPBL{85.0\%}} \\ \hline
            AutoEncoder-VAE & 75\% & 95.1\% & 78.5\% & 57.6\% & 50.2\% & 49.7\% & 53.2\% & 24.8\% & 49.1\% & 78.5\% & 62.3\% & 83.6\% \\
            Q-Former & 75\% & 94.7\% & 77.6\% & 56.8\% & 48.3\% & 48.6\% & 52.9\% & 24.5\% & 48.6\% & 77.8\% & 61.7\% & 83.7\% \\
            PTQ-MinMax (8bit)-FT & 75\% & 95.1\% & 78.5\% & 59.1\% & 50.3\% & 50.7\% & 53.9\% & 25.6\% & 49.3\% & 78.4\% & 62.4\% & 84.4\% \\
            PTQ-P4Q (8bit)-FT & 75\% & 95.4\% & 78.9\% & \OPBL{59.6\%} & 50.8\% & \textbf{51.6\%} & 54.5\% & \textbf{\OPBL{26.3\%}} & \textbf{\OPBL{50.0\%}} & 78.7\% & 62.6\% & \textbf{84.7\%}  \\
            \textbf{Ours} & 75\% & \textbf{\OPBL{96.0\%}} & \textbf{79.6\%} & \textbf{\OPBL{59.9\%}} & \textbf{50.9\%} & \textbf{51.6\%} & \textbf{54.9\%} & \OPBL{26.2\%} & \OPBL{49.9\%} & \textbf{79.1\%} & \textbf{63.2\%} & \textbf{\OPBL{84.9\%}} \\ \hline
            AutoEncoder-VAE & 87.5\% & 93.7\% & 73.8\% & 56.2\% & 48.1\% & 47.3\% & 51.1\% & 23.7\% & 45.9\% & 75.9\% & 60.8\% & 81.7\% \\
            Q-Former & 87.5\% & 93.2\% & 71.3\% & 54.9\% & 47.2\% & 46.4\% & 49.8\% & 22.6\% & 45.8\% & 74.7\% & 60.2\% & 79.9\% \\
            PTQ-MinMax (4bit)-FT & 87.5\% & 84.9\% & 71.3\% & 51.8\% & 43.7\% & 46.3\% & 46.9\% & 19.1\% & 42.4\% & 67.5\% & 55.3\% & 76.8\% \\
            PTQ-P4Q (4bit)-FT & 87.5\% & 85.5\% & 71.8\% & 52.5\% & 44.8\% & 46.8\% & 47.4\% & 20.3\% & 44.1\% & 67.9\% & 55.5\% & 76.8\% \\
            \textbf{Ours} & 87.5\% & \textbf{95.6\%} & \textbf{76.5\%} & \textbf{58.2\%} & \textbf{50.3\%} & \textbf{50.7\%} & \textbf{54.2\%} & \textbf{25.2\%} & \textbf{48.7\%} & \textbf{78.6\%} & \textbf{62.1\%} & \textbf{83.1\%}\\ \hline
            AutoEncoder-VAE & 93.75\% & 90.7\% & 66.1\% & 51.2\% & 43.8\% & 40.2\% & 42.7\% & 15.8\% & 38.4\% & 68.8\% & 55.2\% & 71.7\% \\
            Q-Former & 93.75\% & 89.8\% & 64.5\% & 49.8\% & 44.1\% & 39.7\% & 43.2\% & 16.4\% & 36.5\% & 66.3\% & 56.4\% & 73.2\% \\
            \textbf{Ours} & 93.75\% & \textbf{93.6\%} & \textbf{71.1\%} & \textbf{54.9\%} & \textbf{46.7\%} & \textbf{45.4\%} & \textbf{49.2\%} & \textbf{22.8\%} & \textbf{43.2\%} & \textbf{72.5\%} & \textbf{58.6\%} & \textbf{75.8\%} \\ \hline
	\end{tabular}  }
	\label{tab:classification_compare}
\end{table*}

\textbf{(3) Evaluation on Image Retrieval.}
We evaluate the effectiveness of our compressed features on image retrieval tasks using two datasets: CUB-200-2011 and Cars-196. The performance is measured by Recall at 1 (R@1). Tab. \ref{tab:imageRetr_compare} illustrates that our method consistently outperforms existing feature compression approaches across various compression ratios and different backbones. Our feature at a 50\% compression ratio is also superior to the original, uncompressed features on both datasets. P4Q achieves good performance because it uses training data for calibration, giving it an advantage over other zero-shot methods. 
These results, along with findings from other tasks, demonstrate that our method preserves strong semantic and structural information during compression, making it highly suitable for real-world applications where both efficiency and accuracy are critical.

\begin{table*}[!htbp]
	\small
	\centering
	\setlength\tabcolsep{9pt}
	\setlength{\extrarowheight}{5pt}
	\caption{Evaluation on image retrieval. The ``Baseline'' in the table represents the original feature whose dimension is 1024. The ``FT'' denotes ``Fine-tuning'', which indicates that PTQ methods require training data for accuracy alignment. The blue-marked results indicate performance that matches or surpasses the baseline.}
	\resizebox{.86\textwidth}{!}{
		\begin{tabular}{|c|c|cc|cc|} 
			\hline
            \rowcolor{mygray}\multicolumn{2}{|c|}{ }&\multicolumn{2}{|c|}{\textbf{Baseline model: SigLip 2 (ViT-B/16)}} & \multicolumn{2}{|c|}{\textbf{Baseline model: CN-CLIP (ViT-H/14)}} \\ \hline
            \multicolumn{2}{|c|}{\textbf{Dataset}} & CUB-200-2011  & Cars-196 & CUB-200-2011 & Cars-196  \\ \hline
            Method & CR & R@1 & R@1 & R@1 & R@1 \\ \hline
            Baseline & 0 & 64.8 & 85.7 & 59.1 & 79.8\\ \hline
            AutoEncoder-VAE & 50\% & 64.1 & 85.5 & 58.3 & 79.4 \\
            Q-Former & 50\% & 63.7 & 85.2 & 58.1 & 78.9 \\
            \textbf{Ours} & 50\% & \textbf{\OPBL{65.1}} & \textbf{\OPBL{85.9}} & \textbf{\OPBL{59.5}} & \textbf{\OPBL{80.4}} \\ \hline
            AutoEncoder-VAE & 75\% & 62.9 & 84.9 & 57.8 & 78.4 \\ 
            Q-Former & 75\% & 62.7 & 84.3 & 57.5 & 78.2 \\ 
            PTQ-MinMax (8bit)-FT & 75\% & 64.2 & 84.8 & 58.5 & 79.5 \\ 
            PTQ-P4Q (8bit)-FT & 75\% & 64.4 & 85.1 & 58.7 & 79.7 \\
            \textbf{Ours} & 75\% & \textbf{64.5} & \textbf{85.5} & \textbf{\OPBL{59.1}} & \textbf{\OPBL{79.8}} \\ \hline
            AutoEncoder-VAE & 87.5\% & 60.3 & 82.7 & 55.2 & 75.6\\
            Q-Former & 87.5\% & 59.7 & 81.6 & 54.8 & 74.8 \\
            PTQ-MinMax (4bit)-FT & 87.5\% & 61.2 & 80.9 & 54.7 & 73.5 \\
            PTQ-P4Q (4bit)-FT & 87.5\% & 61.8 & 81.7 & 55.3 & 75.1 \\ 
            \textbf{Ours} & 87.5\% & \textbf{63.4} & \textbf{84.6} & \textbf{57.5} & \textbf{78.2} \\ \hline
            AutoEncoder-VAE & 93.75\% & 56.9 & 80.5 & 51.7 & 71.7 \\
            Q-Former & 93.75\% & 56.6 & 79.9 & 51.3 & 71.5 \\
            \textbf{Ours} & 93.75\% & \textbf{61.7} & \textbf{81.8} & \textbf{55.5} & \textbf{73.3} \\ \hline
	\end{tabular}  }
	\label{tab:imageRetr_compare}
\end{table*}

\begin{figure*}
  \centering
  \includegraphics[width=0.96\linewidth]{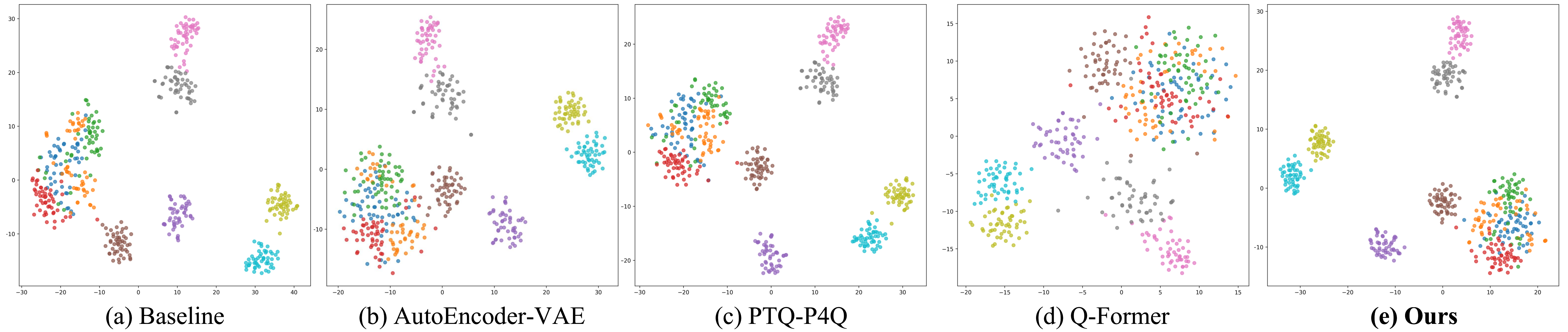}
  \caption{Feature visualization via t-SNE. Each color represents a class sampled on ImageNet. The ``Baseline'' represents the original feature whose dimension is 1024. The feature compression ratio of the compared methods is 75\%. The backbone model is CN-CLIP with ViT-H/14 architecture.}
  \label{fig:visulization} 
\end{figure*}

\textbf{(4) Visualization of Qualitative Evaluation.}
We conduct a qualitative visualization experiment using t-SNE to analyze the feature distributions of compressed representations from different methods. As shown in Fig. \ref{fig:visulization}, the proposed method produces features with clear class separation and tight intra-class clustering, indicating discriminative representation and superior performance. In contrast, VAE and Q-Former yield more scattered and overlapping clusters, suggesting loss of structural information, which correlates with their performance drop in downstream tasks. P4Q shows relatively good clustering, due to its post-quantization calibration using training data, while other methods are zero-shot without fine-tuning. These results demonstrate that our method better preserves class-discriminative structures during compression, leading to superior performance in downstream applications.

\subsection{Ablation Study}\label{sec:ablation}
In this subsection, experiments are conducted to verify the effectiveness of core components and important hyperparameters. Cross-modal retrieval on the Flickr30K-CN dataset is utilized as the task. The feature compression ratio is 50\%. Other settings strictly follow the description in Sec. \ref{sec:settings}. 

\textbf{(1) Effectiveness of Each Component.} 
To evaluate the effectiveness of each component in our framework, we conduct an ablation study by progressively adding modules and measuring their impact on performance. The results are summarized in Tab.~\ref{tab:ablation_component}. We start with a linear compressor, a linear mapping from high-dimensional features to a lower-dimensional space. This serves as a simple form of feature compression. Our proposed ARC module alone already outperforms this linear compressor and VAE, showing the advantage of its compression mechanism. Further improvements are observed when incorporating the MoS and ERGC modules. The MoS module enhances feature representation by refining multiple compressed views through cross-solution attention. Although ERGC is not used during inference, it improves the quality of compressed features during training by preserving the relational structure of the original feature space. 
These results confirm that each component contributes effectively to the overall performance.

\begin{table}
\centering
\setlength\tabcolsep{11pt}
\setlength{\extrarowheight}{3pt}
\caption{Performance of different components in the proposed method. The ``Baseline'' in the table represents the original feature whose dimension is 1024.}\label{tab:ablation_component}
\begin{tabular}{@{}c|ccc@{}}
\hline
Methods & R@1 & R@5 & R@10 \\
\hline
Baseline & 71.2 & 91.4 & 95.5 \\ \hline
Linear compressor & 69.3 & 90.6 & 94.7 \\
AutoEncoder-VAE & 69.8 & 90.6 & 95.0 \\
ARC (Ours) & 69.9 & 90.7 & 95.1 \\
ARC + MoS (Ours) & 70.4 & 91.0 & 95.4 \\
ARC + MoS + ERGC (\textbf{Ours})  & \textbf{71.1}  & \textbf{91.4}  & \textbf{95.6} \\   \hline
\end{tabular}
\end{table}

\textbf{(2) Analysis of the ARC.}
Tab.~\ref{tab:ablation_component} illustrates the effectiveness of ARC. To further analyze the design of ARC, we evaluate the impact of various factors, including the number of decoders used during training, the token length into which the compressed feature is divided, and the mechanism for generating the compressed features. The results are reported in Tab.~\ref{tab:ARC_design}. 

\begin{table*}[!htbp]
\centering
	\setlength\tabcolsep{12pt}
	\setlength{\extrarowheight}{2pt}
    \caption{Ablation study on design of ARC.}\label{tab:ARC_design}
	\resizebox{.96\textwidth}{!}{
    \begin{tabular}{@{}c|cccc|cccc|cc@{}}
    \hline
& \multicolumn{4}{c|}{\textbf{Number of decoders $M$}} & \multicolumn{4}{c|}{\textbf{Token lengths}} & \multicolumn{2}{c}{\textbf{Auto-regressive \textit{v.s}. VAE}}\\ \hline
 & 1 & 3 & \textbf{5} & 7 & 8 & \textbf{16} & 32 & 64 & \textbf{ARC (Auto-regressive)} & VAE with Nested dropout \\
\hline
R@1 & 70.5 & 70.7 & \textbf{71.1} & 70.5 & 70.4 & \textbf{71.1} & 70.7 & 70.6 & \textbf{71.1} & 69.7 \\ 
R@5 & 91.1 & 91.2 & \textbf{91.4} & 91.0 & 90.9 & \textbf{91.4} & 91.2 & 91.2 & \textbf{91.4} & 90.6 \\
R@10 & 95.3 & 95.4 & \textbf{95.6} & 95.3 & 95.2 & \textbf{95.6} & 95.4 & 95.4 & \textbf{95.6} & 95.1 \\  \hline
    \end{tabular}}
\end{table*}

For the number of decoders, the ARC module includes multiple auxiliary decoders during training to enforce better feature reconstruction from the compressed representation. We compare configurations using 1, 3, 5, and 7 decoders. The performance improves as the number increases up to 5, after which it starts to drop. This suggests that too many decoders may introduce conflicting supervision signals or overfitting. Therefore, we choose 5 decoders as the final setting.

For the token lengths, we compare them of 8, 16, 32, and 64. The best performance is achieved with a token length of 16, indicating a good trade-off between local detail preservation and global context modeling.

For the mechanism for generating the compressed features, we compare the auto-regressive (AR) mechanism with VAE under the same backbone. AR generates features iteratively, enabling flexible, arbitrary-ratio compression with progressive refinement. VAE variant uses register tokens and applies nested dropout to ensure an ordered latent representation. This method produces all tokens at once and uses truncation for different ratios. 
Experimental results show that AR outperforms VAE. This is because sequential generation of the AR mechanism captures richer dependencies and allows more precise, controllable feature compression. In contrast, the VAE variant generates all tokens in parallel, which limits its ability to model sequential structure and often leads to less accurate reconstructions. 

\textbf{(3) Analysis of MoS.}
Tab.~\ref{tab:ablation_component} illustrates the effectiveness of MoS. To further analyze the design of MoS, we show how the number of basic solutions affects performance. We test with varying numbers of solutions: 1 (no mixture), 3, 5, 7, and 9. As shown in Tab.~\ref{tab:ablation_MoS}, increasing the number of solutions improves performance up to a point, with 5 solutions achieving the best results. This indicates that combining 5 diverse views strikes a good balance between capturing complementary information and maintaining training stability.

\begin{table}[!htbp]
\centering
	\setlength\tabcolsep{11pt}
	\setlength{\extrarowheight}{2pt}
    \caption{Ablation study on designs of MoS.}\label{tab:ablation_MoS}
    \begin{tabular}{c|ccccc}
    \hline
& \multicolumn{5}{c}{\textbf{Number of solutions $K$}}\\ \hline
 & 1 & 3 & \textbf{5} & 7 & 9\\
\hline
R@1 & 70.2 & 70.4 & \textbf{71.1} & 70.9 & 70.6 \\ 
R@5 & 90.9 & 91.2 & \textbf{91.4} & 91.3 & 91.1 \\
R@10 & 95.1 & 95.3 & \textbf{95.6} & \textbf{95.6} & 95.4 \\  \hline
    \end{tabular}
\end{table}

\textbf{(4) Analysis of ERGC.}
Tab.~\ref{tab:ablation_component} shows that the ERGC improves performance. To further analyze its effect, we compare the relationships (\ie, CLIPScores \cite{hessel2021clipscore}) among entities before and after compression, with and without ERGC, as shown in Tab.~\ref{tab:ablation_ERGC}. When ERGC is applied, the compressed features not only better reflect the original relationships but also form an even more compact and discriminative feature distribution than the original uncompressed features. This suggests that ERGC helps eliminate noise and irrelevant variations while retaining meaningful relational information. The semantic and geometric structure is effectively preserved by ERGC during compression.

\begin{table}
\centering
	\setlength\tabcolsep{8pt}
	\setlength{\extrarowheight}{3pt}
\caption{Cross-modal CLIPScore \cite{hessel2021clipscore} of entity feature relationships before and after compression, with and without ERGC. Higher CLIPScore corresponds to closer relationship.}\label{tab:ablation_ERGC}
\begin{tabular}{@{}c|ccc@{}}
\hline
Compression Ratio & 50\% & 75\% & 87.5\% \\
\hline
Original feature & 31.1 & 31.1 & \textbf{31.1} \\ \hline
Compressed feature w/o ERGC & 31.1 & 30.5 & 30.4 \\
Compressed feature with ERGC (\textbf{Ours}) & \textbf{31.6} & \textbf{31.3} & 31.0  \\   \hline
\end{tabular}
\end{table}

\begin{figure}
  \centering
  \includegraphics[width=0.96\linewidth]{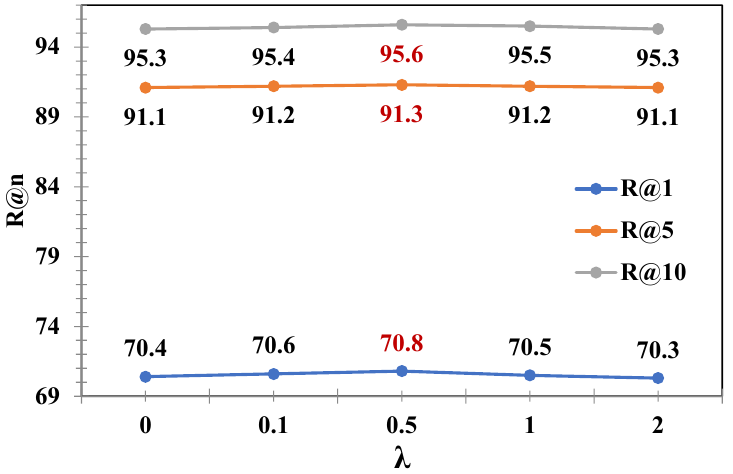}
  \caption{The Recall at n (R@n) curves at different hyperparameters $\lambda$ in Eqns. (\ref{equation:ARC_obj}) and (\ref{equation:MoS_obj}).}
  \label{fig:LambdaTuning} 
\end{figure}

\textbf{(5) Hyperparameter Tuning and Sensitivity.}
We perform a grid search to tune the hyperparameter $\lambda$ in Eqns. (\ref{equation:ARC_obj}) and (\ref{equation:MoS_obj}), which controls the weight of the ERGC loss component in the overall objective function. As depicted in Fig. \ref{fig:LambdaTuning}, the best performance is achieved when $\lambda = 0.5$. We observe that retrieval accuracy, measured by R@n (Recall at n), remains relatively stable across different values of $\lambda$, with fluctuations within 0.5\%. This small variation indicates that our method is not highly sensitive to the choice of $\lambda$, and the performance is robust across a range of settings.

\begin{figure}
  \centering
  \includegraphics[width=0.92\linewidth]{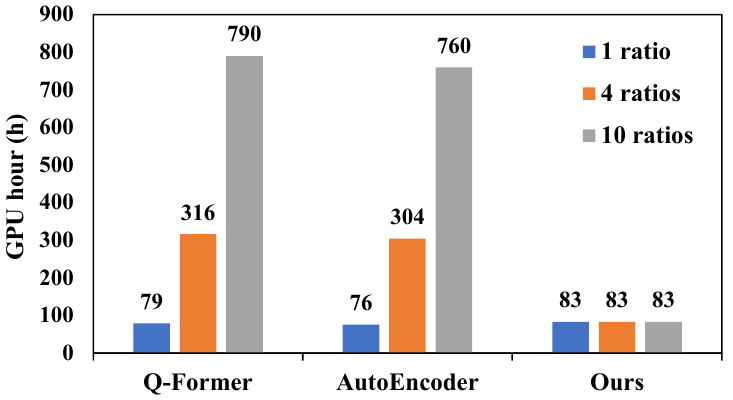}
  \caption{Training time (\ie, GPU hours) comparison with training-based feature compression methods including Q-Former and AutoEncoder, under different numbers of compression ratios (\ie, 1, 4, and 10). Our method requires only a single training run to support all compression ratios.}
  \label{fig:training_efficiency} 
\end{figure}

\textbf{(6) Training Efficiency Analysis.}
Unlike conventional feature compression approaches that require separate training for each compression ratio, our framework only needs a single training run to support ANY compression ratio during inference. This significantly reduces the total training cost and improves practical usability. 
We also compare the GPU hours required for a single training run of our method with those of Q-Former \cite{li2023blip} and Autoencoder \cite{bank2023autoencoders} under the same backbone architecture. As shown in Fig.~\ref{fig:training_efficiency}, our method has a slightly longer training time compared with these baselines. 
However, this small overhead is a one-time cost, regardless of how many compression ratios are used. In contrast, for Q-Former and Autoencoder, training time scales linearly with the number of compression ratios. For example, training 10 compression ratios takes $\sim$790 hours for Q-Former and $\sim$760 hours for Autoencoder, while our method still requires only 83 hours. 
This comparison clearly shows that our method is not only more efficient in total training cost but also more flexible in real-world applications where multiple compression levels may be needed.

\begin{figure}
  \centering
  \includegraphics[width=0.98\linewidth]{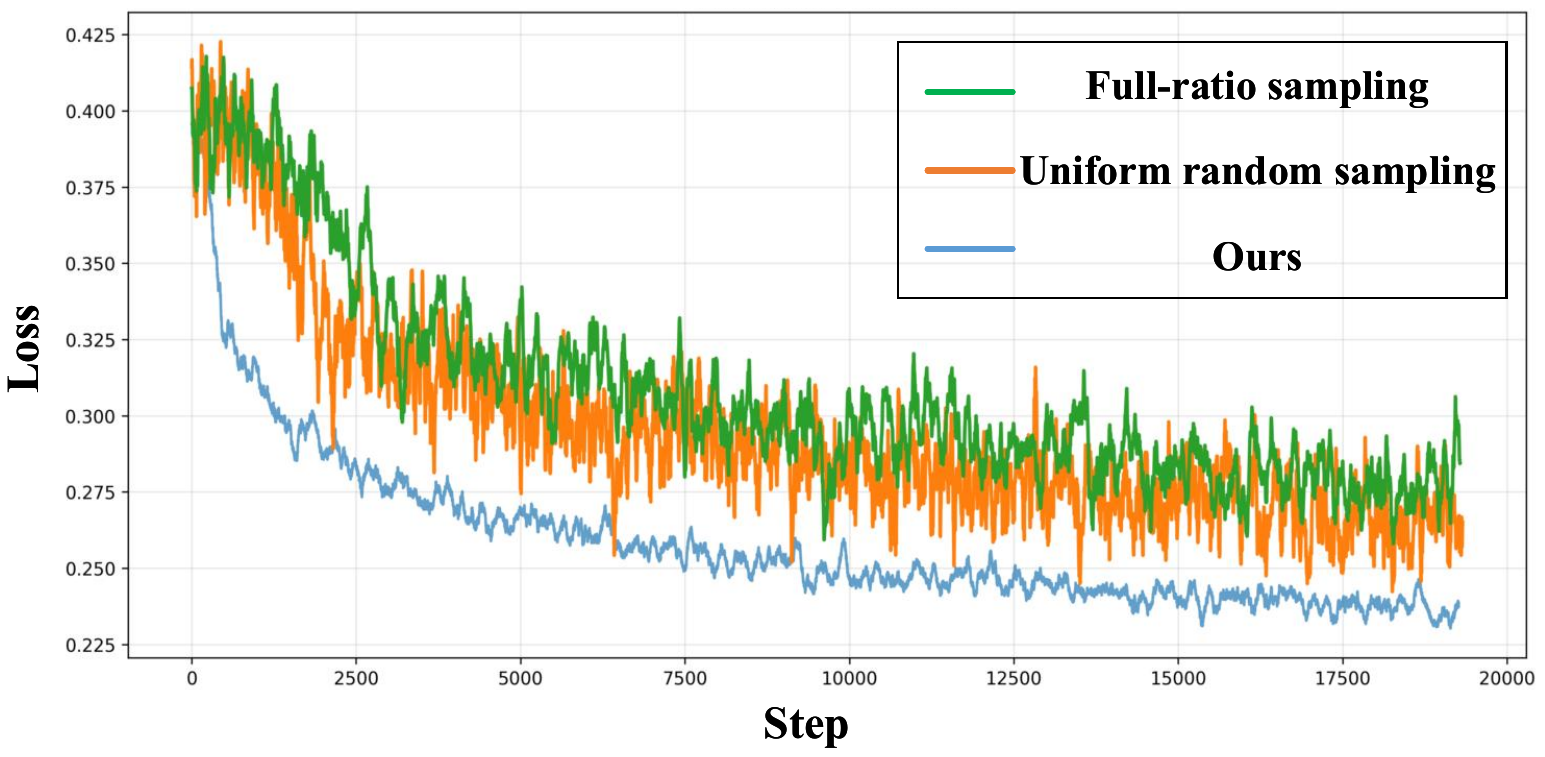}
  \caption{Training loss curves for different training schemes. Note that ``Full-ratio sampling'' uses all compression ratios in each training step; ``uniform random sampling'' selects ratios uniformly at random; ``Ours'' represents our progressive compression training scheme with the dynamic sampling strategy.}
  \label{fig:LossCurve} 
\end{figure}

\textbf{(7) Effectiveness of progressive compression training scheme.} 
We evaluate the effectiveness of our progressive compression training scheme in Fig. \ref{fig:LossCurve}, which compares the training loss curves of three sampling strategies: (i) full-ratio sampling (using all ratios in each step), (ii) uniform random sampling (sampling ratios uniformly at random), and (iii) our dynamic sampling strategy, which uses a Beta-distributed sampling with gradually shifting parameters. The training loss curves show that our approach converges faster and more stably than the alternatives. In contrast, uniform random and full-ratio sampling exhibit slower convergence and higher fluctuation. These results demonstrate that our progressive compression training scheme, by starting with higher compression (coarser reconstruction) and gradually enabling finer details, helps the model learn more effectively and stabilizes training.

\begin{table}[h]
\centering
	\setlength\tabcolsep{11pt}
	\setlength{\extrarowheight}{2pt}
\caption{Evaluation of the dynamic compression ratio adjustment on ImageNet.}\label{tab:dynamic_adjustment}
\begin{tabular}{@{}c|c|c@{}}
\hline
Method & CR & Accuracy \\
\hline
Baseline & all samples: 87.5\% & 58.2\% \\ \hline
\multirow{3}{*}{{  Dynamic CR Adjustment}} & hard samples: 75\% & \multirow{3}*{\textbf{59.1\%}} \\ 
 & easy samples: 93.75\% &  \\ 
 & normal samples: 87.5\% &  \\ 
\hline
\end{tabular}
\end{table}

\textbf{(8) Dynamic Compression Ratio Adjustment.} 
One unique advantage of our method is its ability to dynamically adjust the compression ratio based on available resources or task demands, enabling a flexible trade-off between efficiency and performance.

To evaluate this advantage, we categorize samples into three groups: hard samples: the bottom 15\% classes in terms of accuracy; easy samples: the top 30\% classes in terms of accuracy; normal samples: all remaining samples. We compare two settings: baseline samples: all samples are compressed at a fixed ratio of 87.5\%; dynamic setting: hard samples use a lower compression ratio (75\%), easy samples use a higher ratio (93.75\%), and normal samples remain at 87.5\%. 
As shown in Tab.~\ref{tab:dynamic_adjustment}, this simple adjustment leads to a noticeable improvement in accuracy (\eg, +0.9\%), confirming that increasing the feature dimension for challenging cases helps recover lost discriminative information.
Such dynamic adjustment not only enhances robustness but also provides a practical way to balance computation cost and accuracy in real-world applications with varying resource constraints.

\section{Conclusion}\label{conclusion}
In this paper, we propose an Arbitrary Ratio Feature Compression (ARFC) framework, which enables flexible and efficient feature compression at any desired ratio without retraining. The proposed method consists of three components: Arbitrary Ratio Compressor (ARC),  Mixture of Solutions (MoS), and Entity Relation Graph Constraint (ERGC). The ARC achieves the ARFC via an auto-regressive style model that uses next token prediction to generate features with different compression ratios. MoS refines the compressed features by aggregating multiple basic solutions. ERGC preserves the semantic and structural relationships of the compressed features, further enhancing the representation of the compressed features. 
Extensive experiments on three tasks, including cross-modal retrieval, image classification, and image retrieval, demonstrate that ARFC achieves state-of-the-art performance under various compression settings. 

We foresee three directions for future research in
this area. First, it would be promising to extend the framework to handle more modalities, such as audio and video, which could enable broader applications across different domains. Second, integrating the proposed compression approach into multi-modal large models offers a promising path toward improving their computational efficiency adaptively. Third, we can also investigate online adjustment of compression ratios based on real-time feedback from the execution environment, such as device load or network condition. It could further enhance the system’s adaptability in dynamic environments.

\bibliographystyle{IEEEtran}
\bibliography{sn-bibliography}

\begin{thebibliography}{10}
\providecommand{\url}[1]{#1}
\csname url@samestyle\endcsname
\providecommand{\newblock}{\relax}
\providecommand{\bibinfo}[2]{#2}
\providecommand{\BIBentrySTDinterwordspacing}{\spaceskip=0pt\relax}
\providecommand{\BIBentryALTinterwordstretchfactor}{4}
\providecommand{\BIBentryALTinterwordspacing}{\spaceskip=\fontdimen2\font plus
\BIBentryALTinterwordstretchfactor\fontdimen3\font minus \fontdimen4\font\relax}
\providecommand{\BIBforeignlanguage}[2]{{%
\expandafter\ifx\csname l@#1\endcsname\relax
\typeout{** WARNING: IEEEtran.bst: No hyphenation pattern has been}%
\typeout{** loaded for the language `#1'. Using the pattern for}%
\typeout{** the default language instead.}%
\else
\language=\csname l@#1\endcsname
\fi
#2}}
\providecommand{\BIBdecl}{\relax}
\BIBdecl

\bibitem{ping2013review}
D.~Ping~Tian \emph{et~al.}, ``A review on image feature extraction and representation techniques,'' \emph{International journal of multimedia and ubiquitous engineering}, vol.~8, no.~4, pp. 385--396, 2013.

\bibitem{lu2007survey}
D.~Lu and Q.~Weng, ``A survey of image classification methods and techniques for improving classification performance,'' \emph{International journal of Remote sensing}, vol.~28, no.~5, pp. 823--870, 2007.

\bibitem{wang2016comprehensive}
K.~Wang, Q.~Yin, W.~Wang, S.~Wu, and L.~Wang, ``A comprehensive survey on cross-modal retrieval,'' \emph{arXiv preprint arXiv:1607.06215}, 2016.

\bibitem{saxena2017review}
A.~Saxena, M.~Prasad, A.~Gupta, N.~Bharill, O.~P. Patel, A.~Tiwari, M.~J. Er, W.~Ding, and C.-T. Lin, ``A review of clustering techniques and developments,'' \emph{Neurocomputing}, vol. 267, pp. 664--681, 2017.

\bibitem{radford2021learning}
A.~Radford, J.~W. Kim, C.~Hallacy, A.~Ramesh, G.~Goh, S.~Agarwal, G.~Sastry, A.~Askell, P.~Mishkin, J.~Clark \emph{et~al.}, ``Learning transferable visual models from natural language supervision,'' in \emph{International conference on machine learning}.\hskip 1em plus 0.5em minus 0.4em\relax PmLR, 2021, pp. 8748--8763.

\bibitem{jia2022feature}
W.~Jia, M.~Sun, J.~Lian, and S.~Hou, ``Feature dimensionality reduction: a review,'' \emph{Complex \& Intelligent Systems}, vol.~8, no.~3, pp. 2663--2693, 2022.

\bibitem{dickey1979distribution}
D.~A. Dickey and W.~A. Fuller, ``Distribution of the estimators for autoregressive time series with a unit root,'' \emph{Journal of the American statistical association}, vol.~74, no. 366a, pp. 427--431, 1979.

\bibitem{dubey2021decade}
S.~R. Dubey, ``A decade survey of content based image retrieval using deep learning,'' \emph{IEEE Transactions on Circuits and Systems for Video Technology}, vol.~32, no.~5, pp. 2687--2704, 2021.

\bibitem{rawat2017deep}
W.~Rawat and Z.~Wang, ``Deep convolutional neural networks for image classification: A comprehensive review,'' \emph{Neural computation}, vol.~29, no.~9, pp. 2352--2449, 2017.

\bibitem{li2023blip}
J.~Li, D.~Li, S.~Savarese, and S.~Hoi, ``Blip-2: Bootstrapping language-image pre-training with frozen image encoders and large language models,'' in \emph{International conference on machine learning}.\hskip 1em plus 0.5em minus 0.4em\relax PMLR, 2023, pp. 19\,730--19\,742.

\bibitem{pinheiro2021variational}
L.~Pinheiro~Cinelli, M.~Ara{\'u}jo~Marins, E.~A. Barros~da Silva, and S.~Lima~Netto, ``Variational autoencoder,'' in \emph{Variational methods for machine learning with applications to deep networks}.\hskip 1em plus 0.5em minus 0.4em\relax Springer, 2021, pp. 111--149.

\bibitem{liu2021post}
Z.~Liu, Y.~Wang, K.~Han, W.~Zhang, S.~Ma, and W.~Gao, ``Post-training quantization for vision transformer,'' \emph{Advances in Neural Information Processing Systems}, vol.~34, pp. 28\,092--28\,103, 2021.

\bibitem{liu2022learning}
Y.~Liu, J.~Cao, B.~Li, W.~Hu, and S.~Maybank, ``Learning to explore distillability and sparsability: a joint framework for model compression,'' \emph{IEEE Transactions on Pattern Analysis and Machine Intelligence}, 2022.

\bibitem{liu2019knowledge}
Y.~Liu, J.~Cao, B.~Li, C.~Yuan, W.~Hu, Y.~Li, and Y.~Duan, ``Knowledge distillation via instance relationship graph,'' in \emph{Proceedings of the IEEE/CVF Conference on Computer Vision and Pattern Recognition}, 2019, pp. 7096--7104.

\bibitem{ruan2020edp}
X.~Ruan, Y.~Liu, C.~Yuan, B.~Li, W.~Hu, Y.~Li, and S.~Maybank, ``Edp: An efficient decomposition and pruning scheme for convolutional neural network compression,'' \emph{IEEE Transactions on Neural Networks and Learning Systems}, vol.~32, no.~10, pp. 4499--4513, 2020.

\bibitem{abdi2010principal}
H.~Abdi and L.~J. Williams, ``Principal component analysis,'' \emph{Wiley interdisciplinary reviews: computational statistics}, vol.~2, no.~4, pp. 433--459, 2010.

\bibitem{martinez2001pca}
A.~M. Martinez and A.~C. Kak, ``Pca versus lda,'' \emph{IEEE transactions on pattern analysis and machine intelligence}, vol.~23, no.~2, pp. 228--233, 2001.

\bibitem{maaten2008visualizing}
L.~v.~d. Maaten and G.~Hinton, ``Visualizing data using t-sne,'' \emph{Journal of machine learning research}, vol.~9, no. Nov, pp. 2579--2605, 2008.

\bibitem{tenenbaum2000global}
J.~B. Tenenbaum, V.~d. Silva, and J.~C. Langford, ``A global geometric framework for nonlinear dimensionality reduction,'' \emph{science}, vol. 290, no. 5500, pp. 2319--2323, 2000.

\bibitem{patel2013image}
T.~S. Patel, R.~Modi, and K.~J. Patel, ``Image compression using dwt and vector quantization,'' \emph{International Journal of Innovative Research in Computer and Communication Engineering}, vol.~1, no.~3, p. 653, 2013.

\bibitem{wang2014generalized}
W.~Wang, Y.~Huang, Y.~Wang, and L.~Wang, ``Generalized autoencoder: A neural network framework for dimensionality reduction,'' in \emph{Proceedings of the IEEE conference on computer vision and pattern recognition workshops}, 2014, pp. 490--497.

\bibitem{cheng2018deep}
Z.~Cheng, H.~Sun, M.~Takeuchi, and J.~Katto, ``Deep convolutional autoencoder-based lossy image compression,'' in \emph{2018 Picture Coding Symposium (PCS)}.\hskip 1em plus 0.5em minus 0.4em\relax IEEE, 2018, pp. 253--257.

\bibitem{zhang2018local}
J.~Zhang, J.~Yu, and D.~Tao, ``Local deep-feature alignment for unsupervised dimension reduction,'' \emph{IEEE transactions on image processing}, vol.~27, no.~5, pp. 2420--2432, 2018.

\bibitem{yao2024deco}
L.~Yao, L.~Li, S.~Ren, L.~Wang, Y.~Liu, X.~Sun, and L.~Hou, ``Deco: Decoupling token compression from semantic abstraction in multimodal large language models,'' \emph{arXiv preprint arXiv:2405.20985}, 2024.

\bibitem{zhao2023post}
X.~Zhao, R.~Xu, and X.~Guo, ``Post-training quantization or quantization-aware training? that is the question,'' in \emph{2023 China Semiconductor Technology International Conference (CSTIC)}.\hskip 1em plus 0.5em minus 0.4em\relax IEEE, 2023, pp. 1--3.

\bibitem{schuster1999review}
G.~M. Schuster, G.~Melnikov, and A.~K. Katsaggelos, ``A review of the minimum maximum criterion for optimal bit allocation among dependent quantizers,'' \emph{IEEE Transactions on Multimedia}, vol.~1, no.~1, pp. 3--17, 1999.

\bibitem{wei2022qdrop}
X.~Wei, R.~Gong, Y.~Li, X.~Liu, and F.~Yu, ``Qdrop: Randomly dropping quantization for extremely low-bit post-training quantization,'' \emph{arXiv preprint arXiv:2203.05740}, 2022.

\bibitem{li2021brecq}
Y.~Li, R.~Gong, X.~Tan, Y.~Yang, P.~Hu, Q.~Zhang, F.~Yu, W.~Wang, and S.~Gu, ``Brecq: Pushing the limit of post-training quantization by block reconstruction,'' \emph{arXiv preprint arXiv:2102.05426}, 2021.

\bibitem{yuan2022ptq4vit}
Z.~Yuan, C.~Xue, Y.~Chen, Q.~Wu, and G.~Sun, ``Ptq4vit: Post-training quantization for vision transformers with twin uniform quantization,'' in \emph{European conference on computer vision}.\hskip 1em plus 0.5em minus 0.4em\relax Springer, 2022, pp. 191--207.

\bibitem{hendrycks2016gaussian}
D.~Hendrycks and K.~Gimpel, ``Gaussian error linear units (gelus),'' \emph{arXiv preprint arXiv:1606.08415}, 2016.

\bibitem{lin2021fq}
Y.~Lin, T.~Zhang, P.~Sun, Z.~Li, and S.~Zhou, ``Fq-vit: Post-training quantization for fully quantized vision transformer,'' \emph{arXiv preprint arXiv:2111.13824}, 2021.

\bibitem{sun2024p4q}
H.~Sun, R.~Wang, Y.~Li, X.~Cao, X.~Jiang, Y.~Hu, and B.~Zhang, ``P4q: Learning to prompt for quantization in visual-language models,'' \emph{arXiv preprint arXiv:2409.17634}, 2024.

\bibitem{gawlikowski2023survey}
J.~Gawlikowski, C.~R.~N. Tassi, M.~Ali, J.~Lee, M.~Humt, J.~Feng, A.~Kruspe, R.~Triebel, P.~Jung, R.~Roscher \emph{et~al.}, ``A survey of uncertainty in deep neural networks,'' \emph{Artificial Intelligence Review}, vol.~56, no. Suppl 1, pp. 1513--1589, 2023.

\bibitem{abdar2021review}
M.~Abdar, F.~Pourpanah, S.~Hussain, D.~Rezazadegan, L.~Liu, M.~Ghavamzadeh, P.~Fieguth, X.~Cao, A.~Khosravi, U.~R. Acharya \emph{et~al.}, ``A review of uncertainty quantification in deep learning: Techniques, applications and challenges,'' \emph{Information fusion}, vol.~76, pp. 243--297, 2021.

\bibitem{lan2017fluency}
W.~Lan, X.~Li, and J.~Dong, ``Fluency-guided cross-lingual image captioning,'' in \emph{Proceedings of the 25th ACM international conference on Multimedia}, 2017, pp. 1549--1557.

\bibitem{plummer2015flickr30k}
B.~A. Plummer, L.~Wang, C.~M. Cervantes, J.~C. Caicedo, J.~Hockenmaier, and S.~Lazebnik, ``Flickr30k entities: Collecting region-to-phrase correspondences for richer image-to-sentence models,'' in \emph{Proceedings of the IEEE international conference on computer vision}, 2015, pp. 2641--2649.

\bibitem{krizhevsky2009learning}
A.~Krizhevsky and G.~Hinton, ``Learning multiple layers of features from tiny images,'' 2009.

\bibitem{russakovsky2015imagenet}
O.~Russakovsky, J.~Deng, H.~Su, J.~Krause, S.~Satheesh, S.~Ma, Z.~Huang, A.~Karpathy, A.~Khosla, M.~Bernstein \emph{et~al.}, ``Imagenet large scale visual recognition challenge,'' \emph{International Journal of Computer Vision}, vol. 115, pp. 211--252, 2015.

\bibitem{cimpoi2014describing}
M.~Cimpoi, S.~Maji, I.~Kokkinos, S.~Mohamed, and A.~Vedaldi, ``Describing textures in the wild,'' in \emph{Proceedings of the IEEE conference on computer vision and pattern recognition}, 2014, pp. 3606--3613.

\bibitem{helber2019eurosat}
P.~Helber, B.~Bischke, A.~Dengel, and D.~Borth, ``Eurosat: A novel dataset and deep learning benchmark for land use and land cover classification,'' \emph{IEEE Journal of Selected Topics in Applied Earth Observations and Remote Sensing}, vol.~12, no.~7, pp. 2217--2226, 2019.

\bibitem{goodfellow2013challenges}
I.~J. Goodfellow, D.~Erhan, P.~L. Carrier, A.~Courville, M.~Mirza, B.~Hamner, W.~Cukierski, Y.~Tang, D.~Thaler, D.-H. Lee \emph{et~al.}, ``Challenges in representation learning: A report on three machine learning contests,'' in \emph{International conference on neural information processing}.\hskip 1em plus 0.5em minus 0.4em\relax Springer, 2013, pp. 117--124.

\bibitem{maji2013fine}
S.~Maji, E.~Rahtu, J.~Kannala, M.~Blaschko, and A.~Vedaldi, ``Fine-grained visual classification of aircraft,'' \emph{arXiv preprint arXiv:1306.5151}, 2013.

\bibitem{geiger2013vision}
A.~Geiger, P.~Lenz, C.~Stiller, and R.~Urtasun, ``Vision meets robotics: The kitti dataset,'' \emph{The international journal of robotics research}, vol.~32, no.~11, pp. 1231--1237, 2013.

\bibitem{deng2012mnist}
L.~Deng, ``The mnist database of handwritten digit images for machine learning research,'' \emph{IEEE Signal Processing Magazine}, vol.~29, no.~6, pp. 141--142, 2012.

\bibitem{veeling2018rotation}
B.~S. Veeling, J.~Linmans, J.~Winkens, T.~Cohen, and M.~Welling, ``Rotation equivariant cnns for digital pathology,'' in \emph{International Conference on Medical image computing and computer-assisted intervention}.\hskip 1em plus 0.5em minus 0.4em\relax Springer, 2018, pp. 210--218.

\bibitem{everingham2010pascal}
M.~Everingham, L.~Van~Gool, C.~K. Williams, J.~Winn, and A.~Zisserman, ``The pascal visual object classes (voc) challenge,'' \emph{International journal of computer vision}, vol.~88, no.~2, pp. 303--338, 2010.

\bibitem{wah2011caltech}
C.~Wah, S.~Branson, P.~Welinder, P.~Perona, and S.~Belongie, ``The caltech-ucsd birds-200-2011 dataset,'' 2011.

\bibitem{KrauseStarkDengFei-Fei_3DRR2013}
J.~Krause, M.~Stark, J.~Deng, and L.~Fei-Fei, ``3d object representations for fine-grained categorization,'' in \emph{4th International IEEE Workshop on 3D Representation and Recognition (3dRR-13)}, Sydney, Australia, 2013.

\bibitem{yang2022chinese}
A.~Yang, J.~Pan, J.~Lin, R.~Men, Y.~Zhang, J.~Zhou, and C.~Zhou, ``Chinese clip: Contrastive vision-language pretraining in chinese,'' \emph{arXiv preprint arXiv:2211.01335}, 2022.

\bibitem{tschannen2025siglip}
M.~Tschannen, A.~Gritsenko, X.~Wang, M.~F. Naeem, I.~Alabdulmohsin, N.~Parthasarathy, T.~Evans, L.~Beyer, Y.~Xia, B.~Mustafa \emph{et~al.}, ``Siglip 2: Multilingual vision-language encoders with improved semantic understanding, localization, and dense features,'' \emph{arXiv preprint arXiv:2502.14786}, 2025.

\bibitem{loshchilov2017decoupled}
I.~Loshchilov and F.~Hutter, ``Decoupled weight decay regularization,'' \emph{arXiv preprint arXiv:1711.05101}, 2017.

\bibitem{paszke2019pytorch}
A.~Paszke, S.~Gross, F.~Massa, A.~Lerer, J.~Bradbury, G.~Chanan, T.~Killeen, Z.~Lin, N.~Gimelshein, L.~Antiga \emph{et~al.}, ``Pytorch: An imperative style, high-performance deep learning library,'' \emph{Advances in neural information processing systems}, vol.~32, 2019.

\bibitem{wolf2020transformers}
T.~Wolf, L.~Debut, V.~Sanh, J.~Chaumond, C.~Delangue, A.~Moi, P.~Cistac, T.~Rault, R.~Louf, M.~Funtowicz \emph{et~al.}, ``Transformers: State-of-the-art natural language processing,'' in \emph{Proceedings of the 2020 conference on empirical methods in natural language processing: system demonstrations}, 2020, pp. 38--45.

\bibitem{hessel2021clipscore}
J.~Hessel, A.~Holtzman, M.~Forbes, R.~L. Bras, and Y.~Choi, ``Clipscore: A reference-free evaluation metric for image captioning,'' \emph{arXiv preprint arXiv:2104.08718}, 2021.

\bibitem{bank2023autoencoders}
D.~Bank, N.~Koenigstein, and R.~Giryes, ``Autoencoders,'' \emph{Machine learning for data science handbook: data mining and knowledge discovery handbook}, pp. 353--374, 2023.

\end{thebibliography}

\vfill

\end{document}